\documentclass[10pt,twocolumn,letterpaper]{article}

\usepackage{cvpr}              
\usepackage[dvipsnames, svgnames, x11names]{xcolor}
\usepackage{graphicx}
\usepackage{amsmath}
\usepackage{amssymb}
\usepackage{booktabs}
\usepackage[title]{appendix}
\usepackage{cuted}
\usepackage{marvosym}

\usepackage{multirow}
\usepackage{diagbox}
\usepackage{colortbl}  
\usepackage{array}   
\usepackage{ulem}
\usepackage{cite}

\usepackage[breaklinks,colorlinks]{hyperref}

\begin{document}

\title{Harmonizing output imbalance for defect segmentation on extremely-imbalanced photovoltaic module cells images}

\author{
Jianye Yi\textsuperscript{1,}\thanks{These authors contributed to the work equally and should be regarded as co-first authors.}, Xiaopin Zhong\textsuperscript{1,2,}\footnotemark[1], 
Weixiang Liu\textsuperscript{1,}\textsuperscript{\Letter{}}, Zongze Wu\textsuperscript{1,2}, Yuanlong Deng\textsuperscript{1,3}\\
\textsuperscript{1}Lab. of Machine Vision and Inspection, College of Mechatronics and Control Engineering, \\
Shenzhen University, Nanhai Ave, Shenzhen, 518060, China\\
\textsuperscript{2}Guangdong Artificial Intelligence and Digital Economy Laboratory (Shenzhen),\\ 
Kclian Road, Shenzhen, 518107, China.\\
\textsuperscript{3}Shenzhen Institute of Technology, Jiangjunmao Road, Shenzhen, 518116, China.\\
{\tt\small yijianye2021@email.szu.edu.cn, xzhong@szu.edu.cn, \textsuperscript{\Letter{}}wxliu@szu.edu.cn}\\
{\tt\small zzwu@szu.edu.cn, dengyl@szu.edu.cn}\\
}

\maketitle

\begin{abstract}
The continuous development of the photovoltaic (PV) industry has raised high requirements for the quality of monocrystalline of PV module cells. 
When learning to segment defect regions in PV module cell images, Tiny Hidden Cracks (THC) lead to extremely-imbalanced samples. 
The ratio of defect pixels to normal pixels can be as low as 1:2000. 
This extreme imbalance makes it difficult to segment the THC of PV module cells, which is also a challenge for semantic segmentation.
To address the problem of segmenting defects on extremely-imbalanced THC data, the paper makes contributions from three aspects: 
(1) it proposes an explicit measure for output imbalance;
(2) it generalizes a distribution-based loss that can handle different types of output imbalances; 
and (3) it introduces a compound loss with our adaptive hyperparameter selection algorithm that can keep the consistency of training and inference for harmonizing the output imbalance on extremely-imbalanced input data. 
The proposed method is evaluated on four widely-used deep learning architectures and four datasets with varying degrees of input imbalance. 
The experimental results show that the proposed method outperforms existing methods.

\end{abstract}

\begin{figure}[h]
    \centering
    \includegraphics[width=7cm]{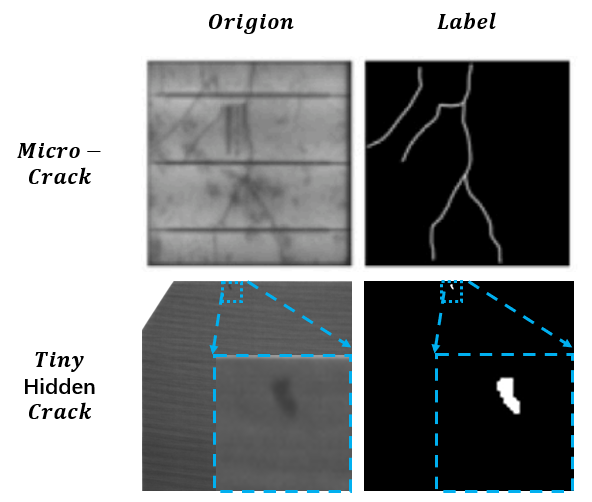}
    \caption{Comparison between tiny hidden cracks and micro-cracks\cite{jiang2022attention} surface defects in monocrystalline PV module cells.} 
    \label{fig:THCsample}
\end{figure}

\section{Introduction}
\label{sec:Introduction}
The growing concern over environmental degradation and energy scarcity has led to a surge in interest in renewable energy sources, particularly solar energy \cite{qian2018micro}.
Photovoltaic (PV) modules, the heart of solar power systems, are devices that convert sunlight directly into electrical energy. 
Broadly speaking, the solar cells in a PV module are categorized into monocrystalline silicon and polycrystalline silicon, based on the material used in their production. 
Owing to the differences in their manufacturing processes, monocrystalline cells typically display a uniform texture, whereas polycrystalline cells exhibit a heterogeneous texture and intricate background due to the presence of numerous randomly shaped and sized lattice particles \cite{chen2020solar}. 
However, solar cells have fragile nature of crystal structure that come into being processing-induced defects such as wafer cracking, breakage, or electrode breakdown; these defects strongly affect the performance, lifetime, and reliability of solar cells \cite{quan2017compressive}. 
Thus, quick and precise evaluations of solar cells during production are required to reproducibly obtain high efficiency and reliable performances in PV module cells.

Tiny Hidden Cracks (THC) of the monocrystalline PV module cells is the most difficult to detect compared with other defects even much more difficult than Micro-crack  \cite{jiang2022attention, xie2023effective} (See Fig.\ref{fig:THCsample}).
For its cracks are extremely fine, but it also has a great impact on the conversion efficiency of monocrystalline PV module cells.
Traditional vision algorithms \cite{liu2019improved, tsai2012defect, yun2009automatic} can detect most of such defects, but due to the minimal contrast, fine contours, and small scale of hidden cracks, their identification remains a challenge for traditional vision algorithms.

In the field of supervised deep learning, surface defects detection methods are divided into four categories: image classification \cite{xie2023effective, jain2022synthetic, liu2020real}, object detection \cite{du2021automated, huang2021texture, ma2022automated}, semantic segmentation \cite{jiang2022attention, niu2023novel, lorena2019fully, tabernik2020segmentation, xi2021ydrsnet, kang2022adaptive}, and image reconstruction based methods \cite{tang2020anomaly, zavrtanik2021draem, pirnay2022inpainting}.
Classification methods are insufficient for accurately localizing surface defects due to their tendency to divide images into regions resulting in a less detailed overall surface defect skeleton, whereas object detection methods suffer from the issue of overlapping bounding boxes and poor performance in localizing surface defects. 
Additionally, image reconstruction methods do not have the capability to effectively capture anomalies in small-scale space, making them inappropriate for detecting extremely small defects. 
Although semantic segmentation labeling costs are high, its pixel-level classification of the input image can accurately locate the position and contour outline of the surface defects, which is of great significance for subsequent work such as defect cause analysis, and is currently the most appropriate method for detecting THC of monocrystalline PV module cells.

\begin{figure}[h]
    \centering
    \includegraphics[width=8cm]{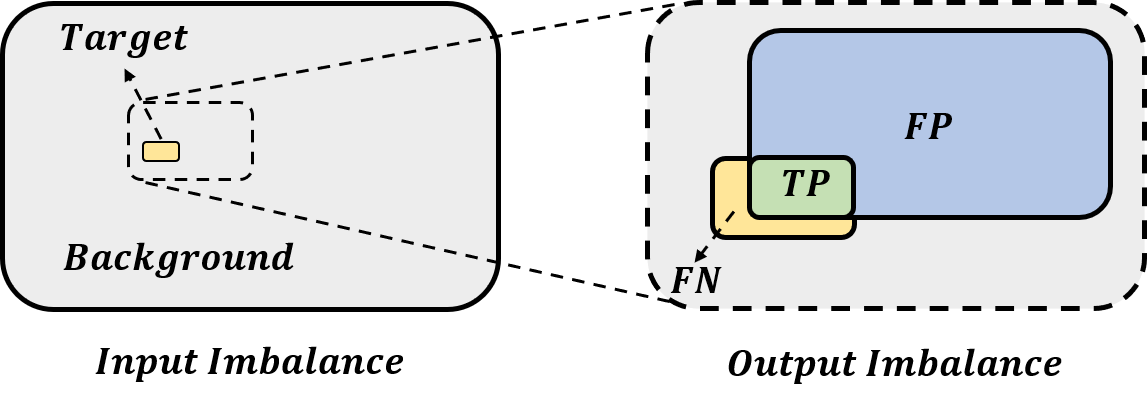}
    
    \caption{The input imbalance for training refers to the class-imbalance among the pixels in the input samples, while the output imbalance for inference refers to the disparity between the false positives ($FP$) and false negatives ($FN$) in the prediction.} 
    \label{fig:imbalance}
\end{figure}

Semantic segmentation, as a high-level task in computer vision, is capable of providing scene understanding at the pixel level by allocating a label to every pixel of an image  \cite{Zhu2016Beyond}.
The application of deep learning to semantic segmentation has witnessed a sharp rise, driven by both academia and industry, demonstrating promising performance capabilities for real-world tasks such as analyzing natural images, industrial surfaces, medical images and autonomous driving \cite{ref_Dung_Autonomous, ref_Li_Automatic, ref_Hesamian_Deep, ref_Hatamizadeh_Unetr, Long2015FCN, ref_Feng_Deep}. 
A comprehensive overview of this trend has recently been reported in \cite{Ulku2022SSDL}.
However, semantic segmentation is also confronted with class imbalance, leading to an unequal distribution of examples among various classes.
One classical example of class imbalance is the foreground-to-background imbalance \cite{Johnson2019Survey}. 
For example, Xi \cite{xi2021ydrsnet} encountered class imbalance when segmenting gear pitting, and they used Dice Loss to replace Cross Entropy Loss to alleviate the impact of this class imbalance.
Recently, Taghanaki \cite{ref_Taghanaki_Combo} defined it as \textit{input imbalance}, stipulating that it refers to the pixel class-imbalance in the input samples, with small target pixels being submerged in a large number of background pixels (as shown in Fig.\ref{fig:imbalance} Left). 
Additionally, they proposed the concept of \textit{output imbalance}, referring to the false positive ($FP$) and false negative ($FN$) imbalance \footnote{For simplicity, $FP$ and $FN$ also denote the numbers of false positive and false negative samples, respectively.} between prediction and ground truth (illustrated in Fig.\ref{fig:imbalance} Right).

In this paper, we address the challenge of semantic segmentation on extremely-imbalanced input data monocrystalline PV module cells\footnote{The dataset is available in  https://github.com/KLIVIS/ DIBE/tree/master}.
The ratio of defect pixels to background pixels is $1:2228$, which is one order of magnitude lower than that of usually used public datasets (as described in Subsection \ref{subsec:Datasets}). This severe class imbalance leads to output imbalance for poor generalization.

Generally speaking, the common solutions for class imbalance can be divided into three levels: data, model and loss function respectively \cite{ref_Gros_Automatic,ref_Christ_Automatic}. 
Most of methods focus on loss design, which only cope with the input imbalance. 
In special applications, output imbalance should be taken into consideration. 
For instance, in the detection of THC, it is desirable not only to maximize the intersection over union ($IoU$) but also to minimize the rate of $FN$ by adjusting the relative size of $FN$ and $FP$.
There are relevant research from the perspective of the loss function to adjust output imbalance.
Kannan \cite{ref_Kannan_Leveraging} selected binary cross-entropy (BCE) loss to expect a more balanced output of $FP$ and $FN$. 
Tversky Loss proposed by Salehi et al. \cite{ref_Salehi_Tversky} applied two hyperparameters to control the weights of $FP$ and $FN$ in the Dice Loss for adjusting the output imbalance. 
However the output imbalance hyperparameters are empirically pre-defined. 
It is not clear to measure the output imbalance and guide the parameter setting.

At present, there are also some open problems with imbalanced input and output in semantic segmentation, the following three issues still exist:

\textbf{Lack of a quantitative measure for output imbalance:} 
The existing method of measuring output imbalance is similar to that of input imbalance, simply using the ratio of $FP$ and $FN$. 
This measurement firstly has a large domain in which the ratio is distributed in $[0, \infty)$. 
It is not convenient to observe the changing trend of output imbalance with visualization. 
Secondly, the influence of true positive ($TP$) on the output imbalance is not considered. 
As an important part of the output confusion matrix, it is unreasonable to ignore $TP$ and only consider $FP$ and $FN$.

\textbf{Lack of ability for adjusting output imbalance in distribution-based loss:}
The diversity of loss functions is important in the context of machine learning.
There are basically two types of loss functions for semantic segmentation, i.e. distribution-based and region-based. 
There are several region-based loss functions for controlling the output imbalance \cite{ref_Salehi_Tversky,ref_Abraham_focaltversky}. 
In distribution-based loss functions, although Taghanak \cite{ref_Taghanaki_Combo} uses weighted binary cross-entropy (WBCE) Loss to adjust the output imbalance, the class weight is actually used to adapt to the input imbalance. 
To our knowledge, there is no work on distribution-based loss that can directly adjust the output imbalance. 
It is even less possible to study the difference between region-based loss and distribution-based loss with regard to the output imbalance.

\textbf{Inconsistency between training and inference for extremely-imbalanced input data: } 
The aim of optimizing the distribution-based loss function is to enhance the pixel-level classification accuracy, which is in line with the objective of attaining pixel accuracy ($PA$).
The objective of region-based loss functions is to maximize the overlap between the predicted target and the labeled target, thereby conforming to the aim of $IoU$.
Although there are some various compound losses to improve the model performance \cite{ref_Taghanaki_Combo,ref_Wong_exponential,ref_Yeung_Unified}, they do not realize that the reason for using a compound loss to elevating the model accuracy is that the optimization objective (loss function) used in training is consistent with the evaluation indicator used for inference \cite{ref_Li_Generalized}.

To address the above problems resulting from extremely-imbalanced input data, we emphasize that \textbf{Data is Imbalanced at Both Ends} (DIBE). 
Especially, in order to harmonize the output imbalance, this paper proposes (i) a new indicator for measuring the output imbalance, (ii) a distribution-based loss function with the ability to harmonize the output imbalance, and (iii) a compound loss adapting to the case of extremely-imbalanced input, as follows:

\textbf{For the absence of output imbalance indicator,} 
we propose a so-called output imbalance index ($OII$) which evaluates output imbalance by taking into account the relative value of $TP$ with respect to $FP+FN$, measured by the $IoU$.
Meanwhile, the most important key of $OII$ is that it not only can be used to monitor the changing of output imbalance during the training process, but also can be used to guide the selection of output imbalance hyperparameters.
To this end, a guidance algorithm is designed, which can effectively reduce the computational complexity and the demand for computing resources to improve the model performance by quickly selecting the hyperparameters.
Please refer to Subsections \ref{subsec:Output_imbalance_index_(OII)} and Appendix \ref{appendixC} for details.

\textbf{For distribution-based loss without adjustment ability of output imbalance,} we propose a distribution-based loss function, DIBE$_{Dis}$ Loss. 
By refining the focal loss \cite{ref_Lin_Focal}, we divide the hard-to-classify samples into two subgroups: $FP$ and $FN$ samples, and then assign different weights to $FP$ and $FN$ samples in the loss respectively. 
So it can harmonize the output imbalance with the distribution-based loss. 
Please refer to Subsections \ref{subsec:Distribution_based_DIBE_Loss} and \ref{subsec:DIBEDis_Loss_can_adjust_output_imbalance} for details.

\begin{figure}[h]
    \centering
    \includegraphics[width=8cm]{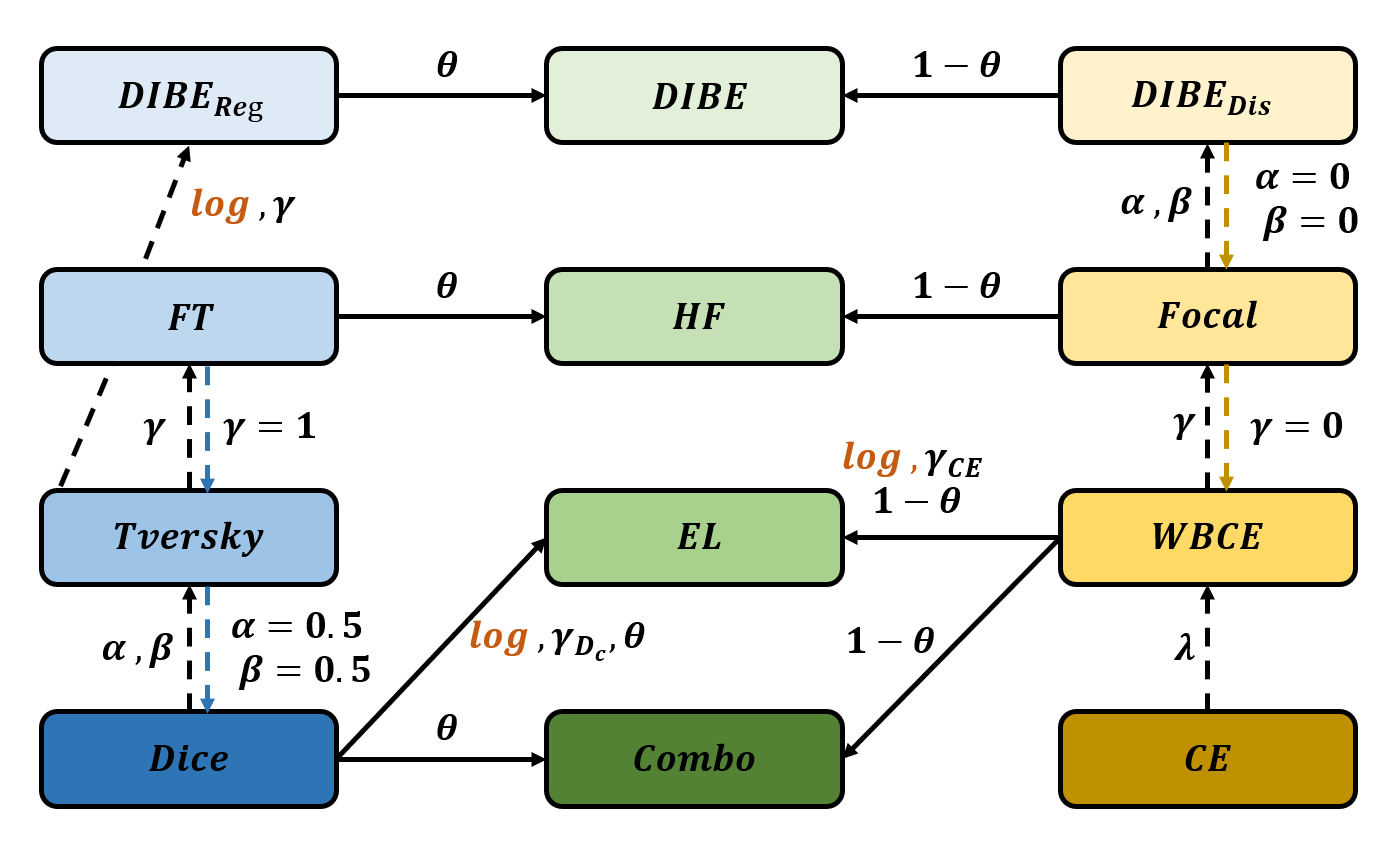}
    
    \caption{The similarities and differences between DIBE Loss and typical losses for semantic segmentation. The $\log$ in figure stands for logarithmic operation. $\theta, \gamma, \lambda, \alpha$ and $\beta$ represent relevant hyperparameters. }
    \label{fig:difLoss}
\end{figure}

\textbf{For inconsistency between training and inference on extremely-imbalanced input data,} 
we propose a new compound loss function, namely DIBE Loss, combining DIBE$_{Dis}$ Loss and DIBE$_{Reg}$ Loss and unifying their hyperparameters (Subsections \ref{subsec:DIBE_Loss},  \ref{subsec:DIBE_Loss_shows_better} and Appendix \ref{appendixB}).
DIBE$_{Dis}$ Loss is the loss function mentioned above, which is more advantageous for optimizing hard-to-classify samples based on Focal loss. 
And DIBE$_{Reg}$ is a region-based loss based on the Tversky Loss, which increases the loss gradient of small targets through a simple logarithmic operation and overcomes the loss over-suppression phenomenon of Focal Tversky (FT) Loss when Tversky coefficient ($Tv$) tends to be $1$. 
Besides, both of them have the capability of harmonizing output imbalance.
The similarities and differences between DIBE Loss and the other loss functions mentioned above can be found in Fig.\ref{fig:difLoss}.

In summary, we mainly harmonize the imbalance at the output end, and provide the following contributions within the framework of DIBE: 
\begin{itemize}
    \item A novel quantitative index $OII$ for measuring the output imbalance is proposed, which satisfies our two Assumptions and is symmetrical, directional, and numerically-normalized; 
    \item For the disability of adjusting output imbalance of the existing distribution-based losses, a new distribution-based loss, DIBE$_{Dis}$ Loss, is generally designed with analogy to region based loss for harmonizing the output imbalance;
    \item In order to keep consistency between training and inference on extremely-imbalanced input data, we propose a new compound loss function, DIBE Loss, for harmonizing training and inference.
\end{itemize}

\section{Related work}
\label{sec:Related_work}
Loss functions in semantic segmentation are crucial. 
A good loss function can make the network learning more efficient. 
Ma \cite{ref_Ma_Loss} and Jadon \cite{ref_Jadon_survey} classified the existing segmentation losses into four categories, namely, distribution-based loss, region-based loss, compound loss and boundary-based Loss. 
For each one, there are many classical loss functions. 
In the following subsections we review their definition, formulation and hyperparameters related to this study.

\subsection{Distribution-based loss}
\label{subsec:Distribution_based_loss}
Cross-entropy \cite{ref_Yi_Automated} is derived from Kullback-Leibler divergence and used to measure the similarity of two distributions for a given random variable or set of events. 
It is often used for classification and works well in segmentation task. 
Binary Cross-Entropy (BCE) Loss is an essential distribution-based loss function for segmentation, and widely used because of its effectiveness. 
For each pixel of image, BCE Loss is defined as:
\begin{equation}
 \mathcal{L}_{BCE} =\left\{
\begin{array}{rcl}
-\log(\hat{p}),       &      & {y = 1}\\
- \log(1-\hat{p}),       &      & {y = 0}
\end{array} \right. ,     
\label{equ:BCE}
\end{equation}
where $y$ denotes the ground truth of pixel sample, and $\hat{p}$ is the predicted probability.

Pihur et al. \cite{ref_Pihur_Weighted}  reported Weighted Binary Cross-Entropy (WBCE) loss function by adding the weight coefficient to measure the importance of different categories based on BCE. 
The pixel categories hard to learn can be effectively trained by increasing the weight. It can be defined as:
\begin{equation}
 \mathcal{L}_{WBCE} =\left\{
\begin{array}{rcl}
-\lambda \log(\hat{p}),       &      & {y = 1}\\
-(1-\lambda)\log(1-\hat{p}),       &      & {y = 0}
\end{array} \right. .     
\label{equ:WBCE}
\end{equation}

Focal Loss is a generalized definition of Cross-Entropy (CE) Loss. 
Lin et al. \cite{ref_Lin_Focal} reduced the loss weight of easily-classified samples by adding an exponential polynomial ${(1 - \hat{p})}^\gamma$, thereby highlighting the loss weight of mis-classified samples. 
This is well known as a good solution of the problem of input imbalance. Its formulation is given as following.
\begin{equation}
 \mathcal{L}_{Focal} =\left\{
\begin{array}{rcl}
-\lambda{(1-\hat{p})}^\gamma \log(\hat{p}),        &      & {y = 1}\\
-(1-\lambda){\hat{p}}^\gamma \log(1-\hat{p}),        &      & {y = 0}
\end{array} \right. ,     
\label{equ:Focal}
\end{equation}
where $\gamma$ represents the hyperparameter used to weight the hard samples. 
In fact, Focal Loss will degenerate into WBCE Loss when $\gamma=0$.
However Focal Loss does not take into account the problem of output imbalance.

\subsection{Region-based loss}
\label{subsec:Region_based_loss}
In the region-based loss category, the most basic one is Dice Loss composed of Dice coefficients proposed by Sudre et al. \cite{ref_Sudre_dice}. 
Dice Loss focuses on the target area, not the background area. 
Therefore, the gradient of Dice Loss is mainly contributed by the target, whereas the gradient from the background is generally small. 
Similar to Focal Loss, Dice Loss is suitable for the tasks with serious input imbalance. 
Nevertheless, it should be emphasized that mis-classification in the target area can have a grave effect on Dice Loss, possibly even leading to stagnation of its loss.
In other words, the training of extremely-imbalanced input data using Dice Loss in segmentation is unstable. 
This shortcoming is fully demonstrated in our experiments: when train networks with our THC dataset, Dice Loss does not decrease rapidly at the beginning, thought its accuracy and segmentation results are better than the distribution-based losses after later convergence. 
Dice Loss and Dice coefficient are formulated as:
\begin{equation}
\begin{array}{rcl}
    \mathcal{L}_{Dice}  = 1-D_c,
\end{array}
    \label{equ:Dice}
\end{equation}
\begin{equation}
\begin{array}{rcl}
    D_c = \dfrac{2TP+\varepsilon}{2TP+FP+FN+\varepsilon}, 
\end{array}
    \label{equ:Dice_coefficient}
\end{equation}
where $TP$, $FP$ and $FN$ refer to the numbers of true positive, false positive and false negative samples respectively. 
$\varepsilon$ is used to prevent the denominator from being zero and increase the numerical stability of loss. Generally, $\varepsilon$ is assigned to $1$.

Dice coefficient uses the same weight for $FP$ samples and $FN$ samples. 
This is obviously unreasonable in the applications pursuing less $FP$ more than less $FN$. 
For the imbalance between $FP$ and $FN$, Salehi et al. \cite{ref_Salehi_Tversky} proposed Tversky coefficient by adding two adjusting parameters $\alpha$ and $\beta$ so as to control the output imbalance. 
Tversky Loss and Tversky coefficient ($Tv$) are defined as follows.
\begin{equation}
\begin{array}{rcl}
    \mathcal{L}_{Tversky} = 1-Tv,
\end{array}
    \label{equ:Tversky}
\end{equation}
\begin{equation}
\begin{array}{rcl}
    Tv = \dfrac{TP+\varepsilon}{TP+\alpha FP+\beta FN+\varepsilon}, 
\end{array}
    \label{equ:Tversky_coefficient}
\end{equation}
where $\alpha$ and $\beta$ are $FP$ and $FN$ regulators respectively. 
It is noted that when $\alpha=\beta=0.5$, $Tv$ degenerates exactly into Dice coefficient.

It has been observed that the Tversky Loss exhibits a constant gradient regardless of the value of $Tv$, meaning that when faced with a small $Tv$, its loss cannot be effectively minimized.
Inspired by Focal Loss, Abraham et al. \cite{ref_Abraham_focaltversky} propose Focal Tversky (FT) Loss, which can employ $\gamma$ to highlight the misclassified samples, thus solving the problem of input imbalance. 
It can be defined as:
\begin{equation}
    \mathcal{L}_{FT} = \big(1-Tv\big)^{\tfrac{1}{\gamma}},  \label{equ:FT}
\end{equation}
where $\gamma$ is another hyperparameter used to tune the contribution of hard samples. FT Loss will degenerate into Tversky Loss if $\gamma=1$.

When $\gamma<1$, FT Loss can effectively pay more attention to hard samples, making their loss contribution greater. 
However, Abraham et al. \cite{ref_Abraham_focaltversky} did not make a more in-depth analysis of its performance on small target segmentation. 
And the loss change is in fact very small when $Tv\rightarrow 1$, leading to over-suppression when the model is close to convergence. They found that $\gamma=0.75$ is the best choice. 

\subsection{Compound Loss}
\label{subsec:Compound_loss}
Inconsistency between the optimization objective (loss function) utilized during training and the evaluation indexes applied in inference can often result in unsatisfactory performance for certain metrics.
The distribution-based loss is constructed based on the classification accuracy of each pixel, which is more consistent with the indicator $PA$.
On the other hand, the region-based losses proposed based on Dice coefficient are more consistent with $IoU$.
However, $IoU$ and $PA$ are both important indicators in semantic segmentation tasks. 
In order to enhance the performance of the model in both aspects, researchers proposed compound losses, which incorporate a combination of distribution-based and region-based losses. 
Furthermore, due to the stability of distribution-based loss during the training process, region-based loss is easier to optimize.
Combo Loss, Exponential Logarithmic (EL) Loss and Hybrid Focal (HF) Loss belong to this kind.

Combo Loss \cite{ref_Taghanaki_Combo} is a basic compound loss function. 
It is defined as a convex combination of CE Loss and Dice Loss,
\begin{equation}
    \mathcal{L}_{Combo} = \dfrac{\theta}{w×h} \sum_{i=1}^{w×h} \mathcal{L}_{WBCE} + (1-\theta)\mathcal{L}_{Dice},  
    \label{equ:Combo}
\end{equation}
where $w$ and $h$ are the width and height of the image. 
$\theta \in(0,1)$ denotes the tuning hyperparameter of the convex combination.
$\mathcal{L}_{WBCE}$ and $\mathcal{L}_{Dice}$ are yielded from Eqs.\ref{equ:WBCE}, \ref{equ:Dice} and \ref{equ:Dice_coefficient}.

The composition form of Exponential Logarithmic (EL) Loss  \cite{ref_Wong_exponential} is similar to that of Combo Loss, but the logarithmic exponent is additionally imposed on Dice coefficient and CE. 
This operation can increase the gradient of small targets in Dice coefficient, and pay more attention to the difficult samples in CE. 
EL Loss can be defined as:
\begin{equation}
    \mathcal{L}_{EL} =  \dfrac{\theta}{w×h} \sum_{i=1}^{w×h} L^{EL}_{CE}  + (1-\theta)\big(-\log(D_c)\big)^{\gamma _{D_c}} , \label{equ:EL}
\end{equation}
where
\begin{equation}
    \mathcal{L}^{EL}_{CE} =\left\{
    \begin{array}{rcl}
    \lambda \big(-\log(\hat{p})\big)^{\gamma _{CE}},       & {y = 1},\\
    (1-\lambda)\big(-\log(1-\hat{p})\big)^{\gamma _{CE}},       & {y = 0},
    \end{array} \right.     \label{equ:EL_CE}
\end{equation}
and $D_c$ is Dice coefficient formulated as in Eqs.\ref{equ:Dice} and \ref{equ:Dice_coefficient}.
$\gamma_{CE}$ is used to tune the attention of CE part for difficult samples. $\gamma_{D_c}$ is used to adjust Dice for different input-imbalance levels.

Different from Combo Loss, in order to better adapt to the highly-imbalanced cases, Yeung et al. \cite{ref_Yeung_Unified} proposed Hybrid Focal (HF) Loss, replacing CE Loss and Dice Loss with Focal Loss and FT Loss respectively. 
This is beneficial to the gradient propagation of small targets. HF Loss $\mathcal{L}_{HF}$ is formulated as:
\begin{equation}
    \mathcal{L}_{HF} = \dfrac{\theta}{w×h} \sum_{i=1}^{w×h} \mathcal{L}_{Focal} + (1-\theta)\mathcal{L}_{FT},  \label{equ:HF}
\end{equation}
where $\mathcal{L}_{Focal}$ and $\mathcal{L}_{FT}$ are given in Eqs.\ref{equ:Focal}, \ref{equ:FT} and \ref{equ:Tversky_coefficient}.

However, HF Loss does not circumvent the inherent shortcomings of Focal Loss and FT Loss, nor does it establish the internal connection between them.

\section{Proposed Method}
\label{sec:Proposed_method}

\subsection{Output imbalance index (OII)}
\label{subsec:Output_imbalance_index_(OII)}
For the class imbalance problem in semantic segmentation, researchers typically use the target-to-background pixel ratio as a metric to measure the severity of input imbalance \cite{ref_Yeung_Unified}. 
However, this ratio may not intuitively and objectively capture the degree of output imbalance. 
This is because, alongside the $FP$ to $FN$ ratio, the number of true positives ($TP$) is also a critical reference.
We believe that the smaller the $TP$ relative to $FP+FN$ is, the more attention should be paid to this imbalance. 
And the simple ratio will also shift greatly depending on which one of $FP$ and $FN$ is selected as the denominator. 
For example, $FP/FN=0.2$ and $FN/FP=5$ show in fact the same degree of output imbalance, but the ratio is totally different.
$FP/FN=0.2$ and $FN/FP=0.2$ denote different direction of output imbalance, but the ratio is completely the same. 
This may lead to ambiguity and is not convenient for objective comparison.

In this study, we propose a new indicator, named as $OII$, to measure the degree of output imbalance, which is symmetrical, directional, and numerical normalized.
$OII$ can measure output imbalance during the training process, and can also be used to intuitively analyze the changes of output imbalance of the model. 
Furthermore, $OII$ can be used to guide the selection of hyperparameters to adjust the output imbalance (see Appendix \ref{appendixC}).




We maintain that, in light of the characteristics of $FP$ and $FN$, the output imbalance index should exhibit a symmetrical, directional, and numerically-normalized pattern.

\textbf{Assumption 1} 
Symmetry in this context refers to a scenario in which the degree of output imbalance represented by $FP$ and $FN$ is invariant when they are interchanged, while directionality is indicative of the relative size between $FP$ and $FN$. 
Specifically, when $FP/FN=0.2$ and $FN/FP=0.2$, the degree of output imbalance is the same, while the relative size between $FP$ and $FN$ and direction of output imbalance are distinct.
Only through the symmetry and directionality, the output imbalance state can be accurately measured, and these two properties can be formulated mathematically in the following way:
\begin{equation}
    f\Big(\dfrac{FP}{FN}\Big)=-f\Big(\dfrac{FN}{FP}\Big),
    \label{equ:definition1}
\end{equation}
where $f(\cdot)$ denotes the index function we aim to identify.

\textbf{Remark 1}  With this assumption, the output imbalance index has the property $ f(x) = -f(\frac {1}{x})$, which is similar to the famous Gini coefficient for long tailed data with imbalance class samples in computer vision \cite{yang2022survey}, and the commonly used economic index measure for income inequality \cite{shorrocks1980class}. 

\textbf{Assumption 2}
Normalization of numerical values means that the value distribution of imbalanced indicators should be within a certain range. 
In combination with \textbf{Assumption 1}, we assume the numerical value of imbalance degree is between $(-1,1)$, which is beneficial to the visualization of numerical values and comparison of the severity of output imbalance. 
Normalization of numerical values can be expressed by the following mathematical formula: 
\begin{equation}
    f\Big(\dfrac{FP}{FN}\Big)  \in (-1, 1).
    \label{equ:definition2}
\end{equation}

\textbf{Remark 2} With this assumption, the output imbalance index has the same property of Pearson coefficient widely used in data analysis \cite{pearson1895vii}.


According to Assumptions 1 and 2 (Eqs.\ref{equ:definition1} and \ref{equ:definition2}), we designed an output imbalance indicator $OII$.
We first define the computation of $x$ according to Eq.\ref{equ:OII_x}, then take the symmetry of $FP$ and $FN$ into consideration, construct $OII$ utilizing $x+\frac{1}{x}$, and restrict its value within the range of $(-1,1)$ (see Eq.\ref{equ:OII}) to guarantee numerical normality. 
Taking into consideration the influence of $TP$ on output imbalance, we propose a regulating factor $k$ (see Eq.\ref{equ:OII_K}) which has a linear correlation with $IoU$ (Eq.\ref{equ:IoU}), thus making $IoU$ a significant factor but not the sole determinant.

\begin{equation}
 OII(x) =\left\{
\begin{array}{rcl}
k\Big(1-\dfrac{1}{x+\frac{1}{x}-1}\Big), & {x \in (0,1]},\\
k\Big(\dfrac{1}{x+\frac{1}{x}-1}-1\Big), & {x \in (1,\infty)},
\end{array} \right.     
\label{equ:OII}
\end{equation}where
\begin{equation}
    x=\frac{FP}{FN},
    \label{equ:OII_x}
\end{equation}and
\begin{equation}
    k=1-\frac{IoU}{2}.
    \label{equ:OII_K}
\end{equation}

\newtheorem{proposition}{Proposition}
\begin{proposition}
$OII$ is symmetrical, directional, and numerical normalized. That is $OII$ (Eqs.\ref{equ:IoU}, \ref{equ:OII_x} and \ref{equ:OII_K}) satisfies constraint condition (Eqs.\ref{equ:definition1} and \ref{equ:definition2}).
\end{proposition}

\begin{figure}
    \centering
    \includegraphics[width=6cm]{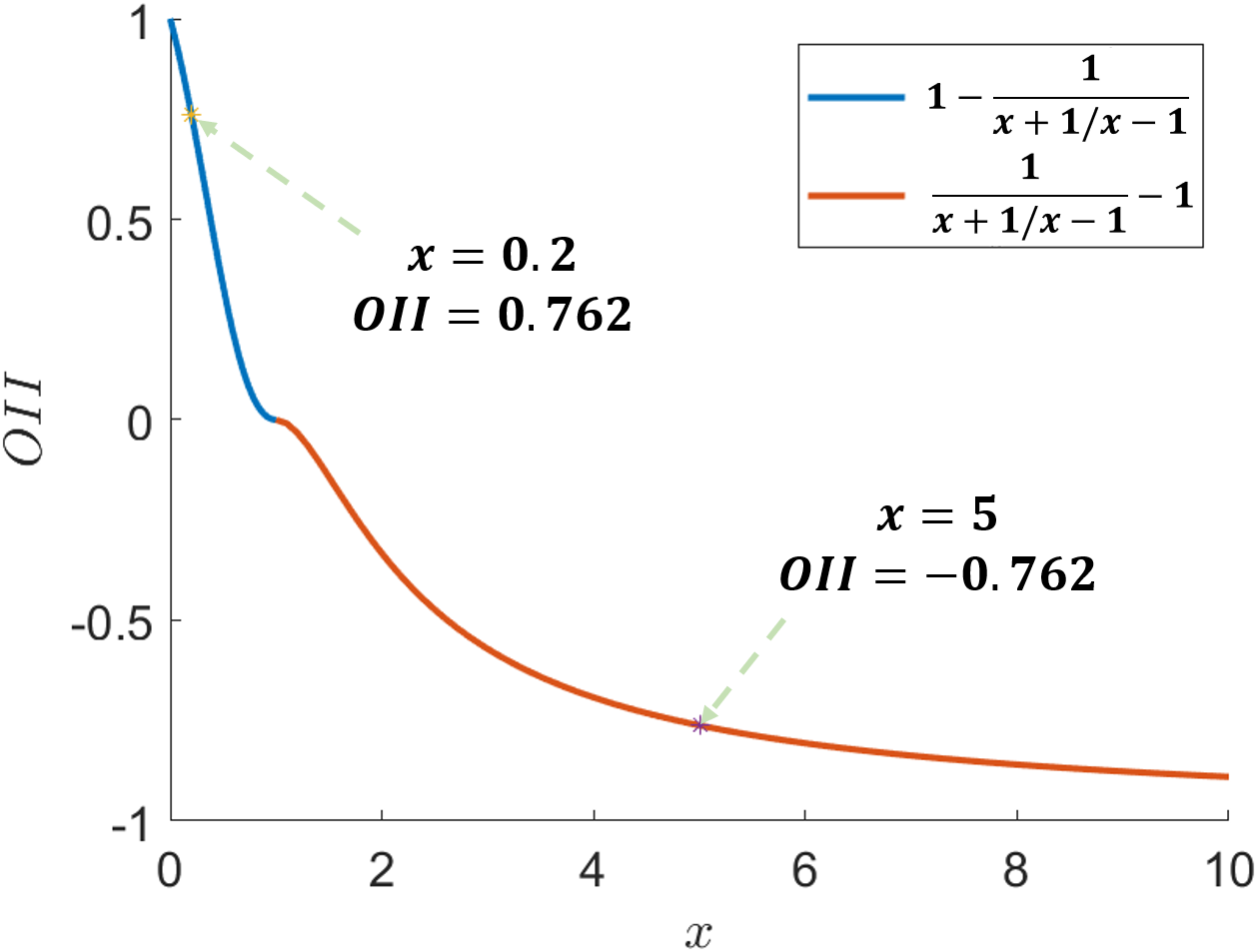}
    \caption{The trend of $OII$ when $k\rightarrow 1$}
  \label{fig:OII}
\end{figure}

The detailed proof of Proposition 1 can be found in Appendix \ref{appendixA}.

\begin{table}\footnotesize
    \centering
    \setlength{\tabcolsep}{2pt}
    \caption{Using $OII$ instead of $FP/FN$ to measure output imbalance.}
    \begin{tabular}{cccccc}
    \toprule  
    No. & TP:FP:FN & FP/FN & IoU & PA & OII\\
    \midrule  
    1 & 2:0.5:0.5 & 1 & 0.667 & 0.800 & 0.000\\
    2 & 2:1:1 & 1 & 0.500 & 0.667 & 0.000\\
    3 & 2:3:3 & 1 & 0.250 & 0.400 & 0.000\\
    4 & 2:5:1 & 5 & 0.250 & 0.286 & -0.667\\
    5 & 2:4:2 & 2 & 0.250 & 0.333 & -0.292\\
    6 & 2:2:4 & 0.5 & 0.250 & 0.500 & 0.292\\
    7 & 2:1:5 & 0.2 & 0.250 & 0.667 & 0.667\\
    8 & 2:1:2 & 0.5 & 0.400 & 0.667 & 0.267\\
    9 & 2:1:0.5 & 2 & 0.571 & 0.667 & -0.238\\
    10 & 2:1:0.2 & 5 & 0.625 & 0.667 & -0.524\\
    \bottomrule 
    \end{tabular}
    \label{table:OII_define}
\end{table}

Fig.\ref{fig:OII} shows the variation of $OII$ with respect to $x$, when $k \rightarrow 1$ ($IoU \rightarrow 0$). 
It can be observed that $OII$ is symmetric with respect to $FP$ and $FN$. 
Moreover, the range of $OII$ lies in the interval of $(-1,1)$. 
Its algebraic sign shows the case of $FP>FN$ or $FP<FN$ and its magnitude denotes the imbalance degree. 
The absolute value of $OII$ close to $0$ suggests a low level of output imbalance, while the magnitude close to $1$ implies a high level.

The value of $k$ is bounded within the range of $[0.5, 1]$, and its value is subject to variation according to changes in the $IoU$.
When the value of $IoU$ is low (i.e. if $TP$ is smaller than $FP+FN$), the degree of output imbalances should be augmented and the value of $k$ tends to be $1$. 
Conversely, the degree of output imbalance should be alleviated and the value of $k$ should tend to $0.5$.

To evaluate the performance of $FP/FN$, $IoU$, $PA$ and $OII$, several representative examples of $FP$, $FN$ and $TP$ are employed, as shown in Table \ref{table:OII_define}.


\subsection{Distribution-based DIBE Loss}
\label{subsec:Distribution_based_DIBE_Loss}
Although Focal Loss can well adapt to the input imbalance, it can not control the output imbalance. 
In fact there is no work reporting a distribution-based loss function directly adjusting the output imbalance. 
\cite{ref_Isensee_Automated} and \cite{ref_Taghanaki_Combo} highlighted that the combination of CE Loss and Dice Loss works better than using them individually in many segmentation tasks. 
The two are based on distribution-based and region-based loss function respectively. 
Therefore, how to better combine distribution-based Loss and region-based Loss is a meaningful work.

Inspired by the idea of region-based Tversky Loss, in this study we propose a distribution-based loss function, called DIBE$_{Dis}$ Loss. 
By adding $FP$ regulator $\alpha$ and $FN$ regulator $\beta$ to Focal Loss, we enable DIBE$_{Dis}$ Loss to adjust the output imbalance. 
The loss function is formulated as: 
\begin{strip} \small
\begin{equation}
 \mathcal{L}_{DIBE_{Dis}} =\left\{
\begin{array}{rcl}
-\lambda(1-\hat{p})^{\gamma}\log(\hat{p}), &    & y=1, \hat{y}=1,\\
-\lambda(1-\hat{p}^{1+\alpha})^{\gamma}(1+\alpha)\log(\hat{p}), &    & y=1, \hat{y}=0,\\
- (1-\lambda)p^{\gamma}\log(1-\hat{p}), &    & y=0, \hat{y}=0,\\
- (1-\lambda)\big(1-(1-\hat{p})^{1+\beta}\big)^{\gamma} (1+\beta)\log(1-\hat{p}), &    &  y=0, \hat{y}=1,
\end{array} \right. 
\label{equ:DIBEDis}
\end{equation}
\begin{equation}
\begin{split}
\left.\frac{\partial \mathcal{L}_{DIBE_{Dis}}}{\partial \hat{p}}\right|_{y=1, \hat{y}=0} = \lambda(1+\alpha)(1-\hat{p}^{1+\alpha})^{\gamma} \left(\gamma(1-\hat{p}^{1+\alpha})^{-1}(1+\alpha){\hat{p}}^{\alpha}\log(\hat{p})-\frac{1}{\hat{p}}\right)
\label{equ:dDIBEDis}
\end{split}.
\end{equation}
\end{strip}
where $\hat{y}$ is predicted label and $y=1, \hat{y}=0$ denotes the regions of $FP$. $\alpha$ and $\beta$ are $FP$ and $FN$ regulator respectively. 
When $\alpha=\beta=0$, the loss will exactly degenerate into Focal Loss.

\begin{figure}[h]
    \centering
    \includegraphics[width=6cm]{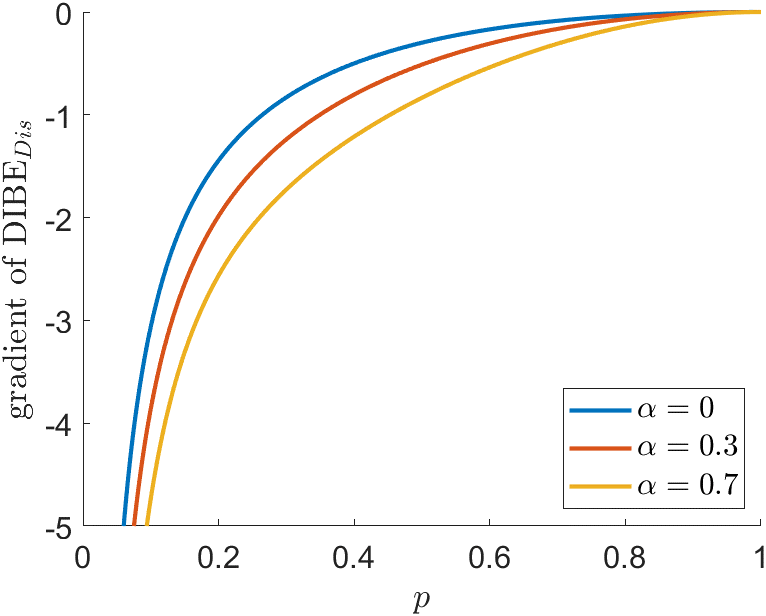}
    \caption{Gradient curve of DIBE$_{Dis}$ under different $\alpha$.}
    \label{fig:Gradient_DIBEDis}
\end{figure}

To analyze the regulation principle of DIBE$_{Dis}$ on output imbalance, the effect of $\alpha$ (likewise for $\beta$) on output imbalance is taken as an example for analysis. 
The derivative of DIBE$_{Dis}$ Loss at $y=1, \hat{y}=0$ is seen as Eq.\ref{equ:dDIBEDis}. 
In order to facilitate comparison and analysis, we fix $\lambda$ and $\gamma$ of DIBE$_{Dis}$ Loss with $0.25$ and $2$ respectively, calculate the derivative with $\alpha\in\{0, 0.3, 0.7\}$ and then get the derivative versus $\hat{y}$ curve shown in Fig.\ref{fig:Gradient_DIBEDis}. 
When $\alpha=0$, DIBE$_{Dis}$ Loss degenerates into Focal Loss. 
It can be observed from the figure that the gradient magnitude of DIBE$_{Dis}$ is increasing as $\alpha$ increases. 
This implies that increasing $\alpha$ can increase the network's attention to $FP$.
In the same way, increasing $\beta$ can increase the network's attention to $FN$.
If $\alpha=1-\beta$ is imposed and $\alpha \neq \beta$, the loss gradient of $FP$ and $FN$ will be different with each other. 
This difference is the key to control the output imbalance, which is how DIBE$_{Dis}$ Loss works. 
For example, when $\alpha=0.3$ and $\beta=0.7$, the loss gradient of $A_{FN}$ (FN area) is always greater than that of $A_{FP}$ (FP area). 
At the moment the loss of $A_{FN}$ contributes more and the network is more inclined to optimize based on $A_{FN}$, making $FN$ decrease more strongly than $FP$. 
Consequently it achieves the purpose of adjusting $FN$ and $FP$, leading to $FN<FP$.

\subsection{DIBE Loss}
\label{subsec:DIBE_Loss}
Combo Loss, EL Loss and HF Loss combine the advantages of distribution-based loss and region-based loss. 
They can not only leverage Dice or Tversky coefficients to deter model parameters from being held at bad local minimum, 
but also gradually learn better model parameters by decreasing $FP$ and $FN$ using a cross entropy term. 
And this combination makes training objectives and inference indexes consistent, finally making the inference performance better. 
Among them, HF Loss proposed by Yeung et al. \cite{ref_Yeung_Unified} is a simple convex combination of Focal Loss and FT Loss, unifying the hyperparameters and demonstrating that HF Loss is better than Focal Loss or FT Loss alone. 
This allows HF Loss to be used for highly-imbalanced input data, and also to adjust the output imbalance. 
However, it does not essentially explore the relationship between the two ends.

In this study we proposed DIBE$_{Dis}$ Loss in Subsection \ref{subsec:Distribution_based_DIBE_Loss} by adding $FP$ and $FN$ regulators $\alpha$ and $\beta$, so that it can not only work on input imbalance but also harmonise output imbalance. 
In the region-based loss, we refers to the region-based part of EL Loss and proposes region-based loss for DIBE, named as DIBE$_{Reg}$ Loss. DIBE$_{Reg}$ Loss can solve the problem that FT Loss will lead to over-suppression of loss then challenge the convergence of model training when $Tv\rightarrow 1$. 
The comparision and gradient analysis of DIBE$_{Reg}$ Loss and FT Loss are seen as Appendix \ref{appendixB}. 
It is in fact a logarithmic form of Tversky Loss with formula as following:
\begin{equation}
    \mathcal{L}_{DIBE_{Reg}} = \big(-\log(Tv)\big)^{\tfrac{1}{\gamma}},  \label{equ:DIBERe}
\end{equation}
where $Tv$ is Tversky coefficient as defined in Eq.\ref{equ:Tversky_coefficient}, and $\gamma$ is our hyperparameter to specify the level of input imbalance.

Then, a compound loss combining DIBE$_{Dis}$ and DIBE$_{Reg}$, namely DIBE Loss, is proposed for extremely-imbalanced input and output. 
The DIBE Loss and HF Loss have an equivalent number of hyperparameters, and the former does not include any extra ones. 
The distinguishing factor between them is that DIBE can regulate output imbalance by DIBE$_{Dis}$, whereas HF lacks this capability. 
Furthermore, DIBE is specifically designed to accommodate extreme input imbalances based on DIBE$_{Reg}$ while HF is not adapted to do so.
It is in fact a unified form with clearer parameter meaning as defined below.
\begin{equation}
    \mathcal{L}_{DIBE} = \dfrac{\theta}{w×h} \sum_{i=1}^{w×h} \mathcal{L}_{DIBE_{Dis}}  + (1-\theta) \mathcal{L}_{DIBE_{Reg}},
    \label{equ:DIBE}
\end{equation}
where $\mathcal{L}_{DIBE_{Dis}}$ and $\mathcal{L}_{DIBE_{Reg}}$ are already given in Eqs.\ref{equ:DIBEDis} and \ref{equ:DIBERe}. 
$\theta$, $\lambda$, $\gamma$ and $\alpha$ in $\mathcal{L}_{DIBE_{Dis}}$ and $\mathcal{L}_{DIBE_{Reg}}$ are the same hyperparameters. 
Herein, $\theta$ is used to tune the combining weight of DIBE$_{Dis}$ Loss and DIBE$_{Reg}$ Loss and we usually set $\theta=0.5$. 
$\lambda$ is the category weight coefficient controlling the input imbalance. 
Generally, the selection of $\lambda$ is based on the ratio of target to background pixels. 
$\gamma$ is also the hyperparameter for proper adjustment considering the degree of input imbalance. 
We set $\gamma\in[0,4]$ in our experiments. 
$\alpha\in[0,1]$ is the hyperparameter to control output imbalance. 
The closer $\alpha$ is to $1$, the stronger the regulation of $FP$ and the weaker the regulation of $FN$ are.

\section{Experiment}
\label{sec:Experiment}
\subsection{Datasets}
\label{subsec:Datasets}
We first describe the collection and preprocessing of monocrystalline PV module cells data, then present the private dataset utilized in the ultimate experiment and finally, divulge the public dataset employed in the experiment.

\begin{figure}
    \centering
    \includegraphics[width=6.5cm]{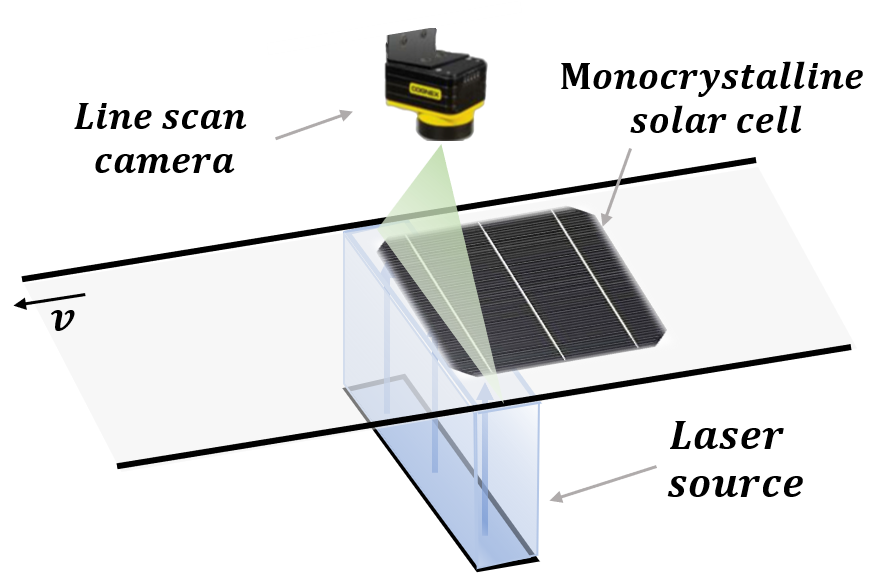}
    \caption{The monocrystalline PV module cells undergo a laser penetration imaging process which involves placing the cells on a conveyor belt and sending them through the laser projection site to capture high-resolution raw images with a line scan camera.}
  \label{fig:product}
\end{figure}

Mono-Crystalline Ingot is cut into monocrystalline PV module wafers using a wire saw, which are subsequently processed into monocrystalline PV module cells by slicing into smaller pieces \cite{abdelhamid2013review}.
Various surface defects are easily produced in this process, among which the tiny hidden cracks are the focus of this study and the most difficult to identify.
To minimize economic losses, monocrystalline PV module cells with tiny hidden cracks can be intercepted and returned to the supplier prior to the commencement of the optical coating process.
Therefore, the PV module cells' images obtained in this study are all images before optical coating process.
Special imaging techniques are used to detect and analyze the tiny hidden cracks defect, which is hard to observe by eyes.
The infrared imaging scheme, commonly used in the market, not only possess a high equipment price but also poor imaging results. 
In order to meet industrial production demands while keeping costs low and achieving good imaging results, this study opted for a laser penetration scheme. 
Fig.\ref{fig:product} illustrates the imaging process of this scheme, where cells are penetrated by laser and scanned with a line scan camera.

\begin{figure}[h]
    \centering
    \includegraphics[width=8cm]{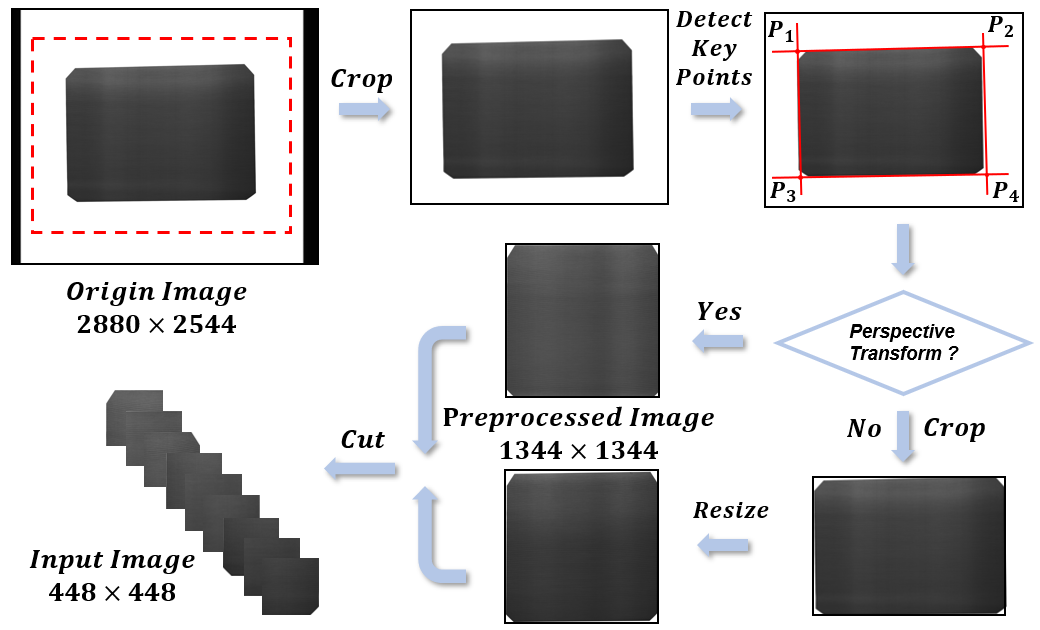}
    \caption{Algorithm Flowchart from High-Resolution Raw Image to $448 \times 448$ Semantic Segmentation Model Input Image.}
    \label{fig:preprocessing}
\end{figure}

Due to the large size of these images, preprocessing was necessary to facilitate efficient network training while avoiding excessive computational resource consumption.
Specifically, the raw $1344 \times 1344$ image is divided into 9 smaller images with a resolution of $448 \times 448$ for each, as illustrated in Fig.\ref{fig:preprocessing}.

\begin{figure*}
  \centering
  \begin{subfigure}{0.13\linewidth}
    \centering
    \includegraphics[width=2.5cm]{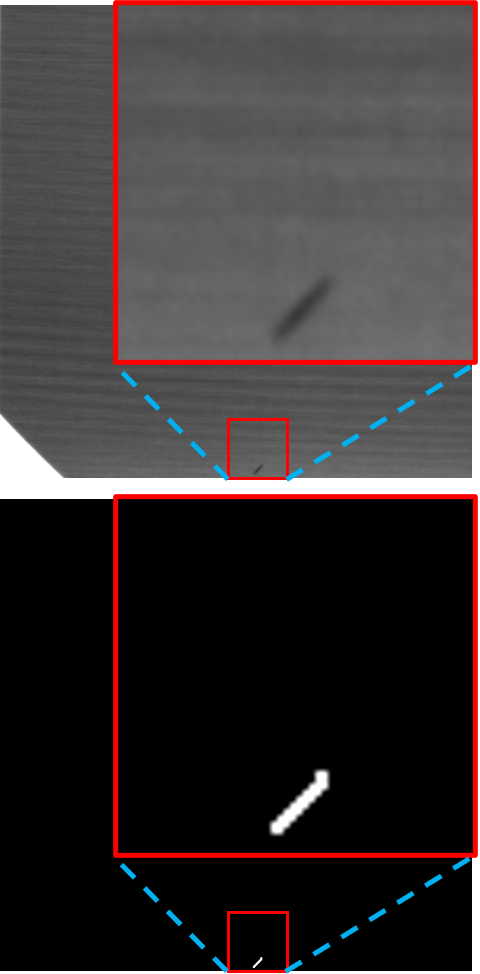}
    \caption{ }
    \label{fig:THC448}
  \end{subfigure}
  \hfill
  \begin{subfigure}{0.13\linewidth}
    \centering
    \includegraphics[width=2.5cm]{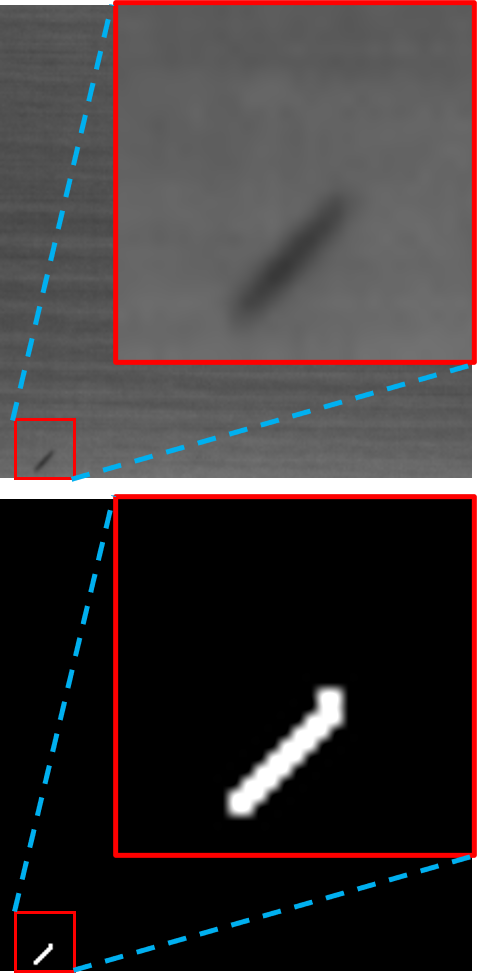}
    \caption{ }
    \label{fig:THC224}
  \end{subfigure}
  \hfill
  \begin{subfigure}{0.13\linewidth}
    \centering
    \includegraphics[width=2.5cm]{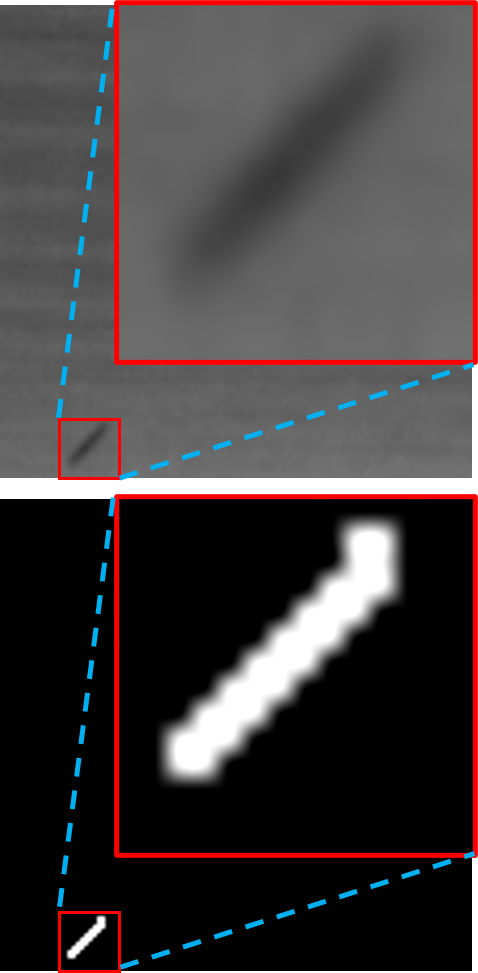}
    \caption{ }
    \label{fig:THC112}
  \end{subfigure}
  \hfill
  \begin{subfigure}{0.13\linewidth}
    \centering
    \includegraphics[width=2.5cm]{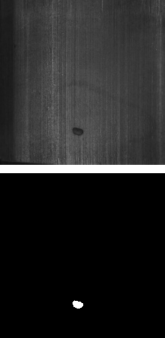}
    \caption{ }
    \label{fig:Blowhole}
  \end{subfigure}
  \hfill
  \begin{subfigure}{0.13\linewidth}
    \centering
    \includegraphics[width=2.5cm]{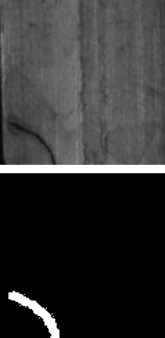}
    \caption{ }
    \label{fig:Crack}
  \end{subfigure}
  \hfill
  \begin{subfigure}{0.13\linewidth}
    \centering
    \includegraphics[width=2.5cm]{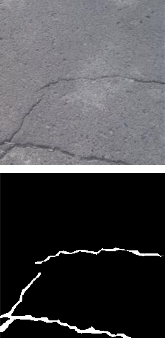}
    \caption{ }
    \label{fig:CFD}
  \end{subfigure}
  \hfill
  \caption{Sample images of our datasets, MTSDD datasets and CFD dataset. Column (a) , (b), (c), (d), (e) and (f) denote pairs of sample images from our THC448, our THC224, our THC112, Blowhole of MTSDD, Crack of MTSDD and CFD respectively. The above of each column is the original image, and the following is the annotation.}
  \label{fig:dataset}
\end{figure*}

In this experiment, a total of $1874$ pairs of input and labelled images of tiny hidden cracks were preprocessed to obtain input images with a resolution of 448x448.
Fig.\ref{fig:THC448} shows some examples of these images. 
Subsequently, we created three samples datasets, namely, THC448, THC224 and THC112, with the resolutions of $448 \times 448$, $224 \times 224$ and $112 \times 112$, respectively.
The respective proportions of pairs used in the training dataset and validation dataset for each of the three datasets were $981$ to $893$, $1053$ to $954$ and $1334$ to $1177$.

\begin{table}\footnotesize
    \centering
    \setlength{\tabcolsep}{3pt}
    \caption{Comparison on datasets selected for experiments. } 
    \begin{tabular}{cccccc}
    \toprule  
    Dataset & train & val & total & FG:BG & \begin{tabular}[c]{@{}c@{}}imbalance\\ level\end{tabular}  \\
    \midrule  
    THC448 & 981 & 893 & 1874 & 1:2228 & extreme\\
    THC224 & 1053 & 954 & 2007 & 1:584 & high\\
    THC112 & 1334 & 1177 & 2511 & 1:191 & medium\\
    Blowhole & 101 & 70 & 171 & 1:333 & high\\
    Crack & 46 & 31 & 77 & 1:113 & medium\\
    CFD & 360 & 348 & 708 & 1:25 & low\\
    \bottomrule 
    \end{tabular}
    \label{table:Datasets}
\end{table}

The experiments are performed on datasets covering the extremely, highly, medium and low imbalanced cases, i.e. our THC448, THC224, THC112 datasets as well as the public datasets Magnetic Tile Surface Defect Dataset\cite{ref_Huang_Surface} (MTSDD) and CrackForest Dataset \cite{ref_Shi_Automatic} (CFD). 
See Table \ref{table:Datasets} for dataset comparison.

The public dataset MTSDD is divided into six sub datasets, also with pixel-level annotations. 
According to the defect type, they are named respectively Blowhole, Crack, Fray, Break, Uneven and Free (no defects). 
Dataset CFD contains shadows, oil stains, water stains and other noises samples of urban road surface. 
Take into account for different levels of imbalance, we select Blowhole, Crack and CFD as well for comparative experiments, see Table \ref{table:Datasets} for the pixel ratio of target to background. 
In order to adapt to the input interface of the semantic segmentation model, we further crop and preprocess the samples in Bloowhole, Crack and CFD. 
Some representative samples are shown in Figs.\ref{fig:Blowhole} \ref{fig:Crack} and \ref{fig:CFD}.

\subsection{Evaluation metrics}
\label{subsec:Evaluation_metrics}
The experiments uses pixel-level performance indicators to evaluate the models and methods. 
The first selected indicator is the intersection over union ($IoU$) between the prediction and the label. 
The larger $IoU$, the higher the overlap of the region between the prediction and the label, implying the better the semantic segmentation performance. 
The second indicator is the pixel accuracy ($PA$), measuring the pixel-wise classification accuracy of the semantic segmentation model. 
The two indicators are defined as follows. 
\begin{equation}
    IoU = \dfrac{TP}{TP+FP+FN},
    \label{equ:IoU}
\end{equation}
\begin{equation}
    PA = \dfrac{TP}{TP+FP}.
    \label{equ:PA}
\end{equation}

\subsection{Model and configuration}
\label{subsec:Model_and_configuration}

\subsubsection{Segmentation model}
\label{subsubsec:Segmentation_model}
The experiments mainly employ several classical semantic segmentation models, such as SegNet \cite{ref_Badrinarayanan_Segnet} using the max-pooling index as a skip connection for upsampling, U-Net \cite{ref_Ronneberger_Unet} applying linear interpolation in the process of upsampling, PSPNet \cite{ref_Zhao_Pyramid} using pyramid pooling module to collect levels of information, and DeepLabv3 \cite{ref_Chen_Rethinking} employing atrous convolution in cascade or in parallel to capture multi-scale context by adopting multiple atrous rates.

\subsubsection{Hyperparameters of loss function}
\label{subsubsec:Hyperparameters_of_Loss_Function}
In the experiment, for the sake of fairness, we basically follow the optimal hyperparameters mentioned in the original references, and set the same hyperparameters in DIBE Loss. 
The basic hyperparameters in all losses are set as:
$\lambda=(0.25,0.75)$, $\theta=0.5$, $(\alpha, \beta)=(0.3, 0.7)$, $\gamma_ {FT}=0.75$, $\gamma_{DIBE_{Reg}}=1.5$, $\gamma_{Focal}=\gamma_{DIBE_{Dis}}=\gamma_{DIBE}=2$, $\gamma_{CE}=\gamma_{D_c}=1$.
In particular, as presented in Table \ref{table:U-Net_dif_Loss}, we obtain the optimal hyperparameters $\lambda$, $\theta$, $\alpha$ and $\gamma$ for each loss. 
Subsequently, in Subsections \ref{subsec:OII_can_reflect_and_guide_output_imbalance} and \ref{subsec:DIBEDis_Loss_can_adjust_output_imbalance}, to investigate how DIBE${Dis}$, DIBE${Reg}$, and Tversky Loss balance the output imbalance, we only fine-tune the hyperparameters $\alpha$ and $\beta$.

\subsubsection{Training configuration}
\label{subsubsec:Training_strategy}
For hardware, a Linux station composed of I7 12700K CPU and two 24G RTX3090Ti GPUs is used in this research for training. 
Pytorch is used as framework. We use stochastic gradient descent as the optimizer with the initial learning rate $0.01$, the momentum $0.9$ and batch size $32$. The experiments set a fixed random seed, and all training datasets are augmented (flip, contrast, brightness, saturation). 
The models are trained $50$ epochs, and every $5$ epochs the validation datasets is tested and $IoU$, $PA$ and $OII$ are estimated. 
All following experiments are performed with the same configuration.

\subsection{$OII$ can better measure output imbalance}
\label{subsec:OII_can_reflect_and_guide_output_imbalance}
\begin{figure}
  \centering
  \includegraphics[width=6cm]{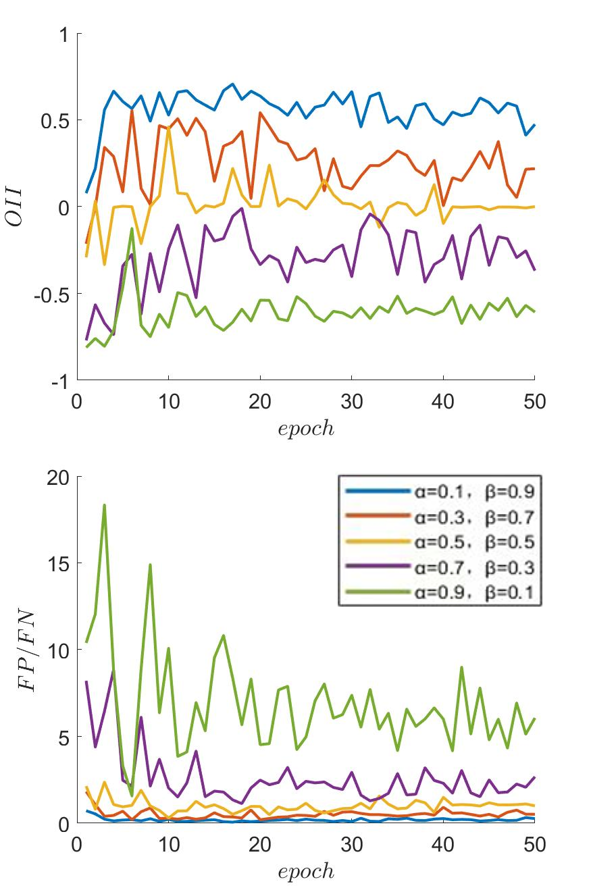}
  \caption{Comparison between $OII$ (top) and traditional output imbalance indicator $FP/FN$ (bottom) under the same data.}
  \label{fig:OIIaFP2FN}
\end{figure}

By leveraging U-Net and training the THC448 dataset with $DIBE_{Reg}$ Loss, this study has visually depicted the curves of $OII$ and $FP/FN$ during the training process, as illustrated in Fig.\ref{fig:OIIaFP2FN}. 
It can be observed that the $FP/FN$ curves are distributed in $(0, 20)$ with an upper limit of infinity, which makes it difficult to measure the influence of imbalances on a same scale. 
For example, when $\alpha=0.1,0.3,0.5$, the curves of $FP/FN$ display similar patterns. 
In contrast, when $OII$ is used as a measure of output imbalance, no such phenomenon exists. 
It is apparent in the chart that the values of $OII$ are dispersed in $(-1, 1)$, thereby making the variation in imbalance quantifiable and easily discernible. 
Furthermore, the proportion of $FP$ and $FN$ can also be inferred from the algebraic sign.

\begin{figure*}[h]
    \centering
    \includegraphics[width=13cm]{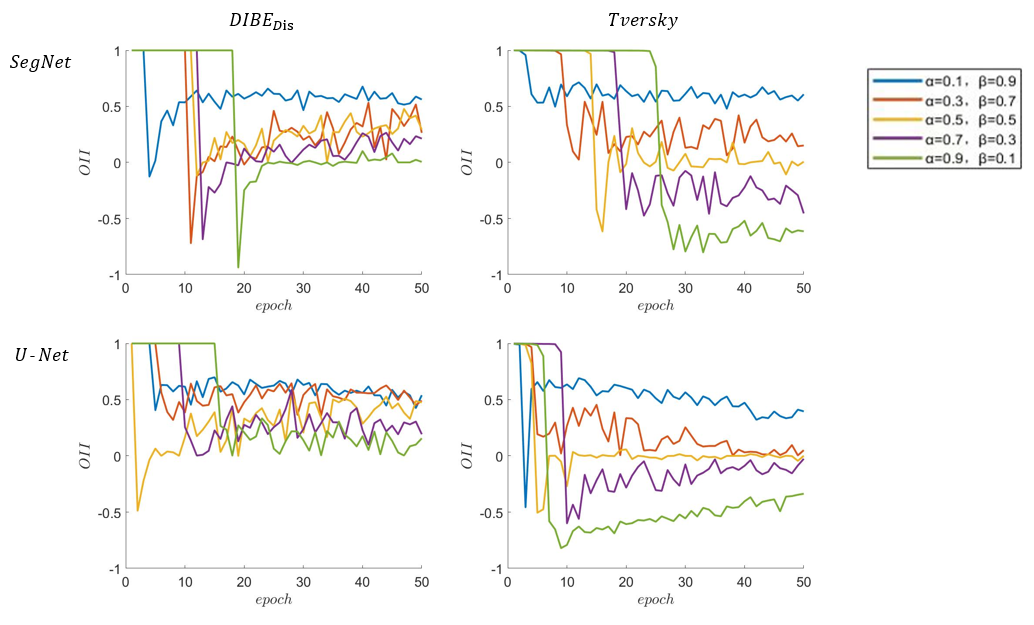}
    
    \caption{Regulation of output imbalance of $\alpha$ in THC448 with different models and different losses. Where the upper row is SegNet and the lower one is U-Net.}
    \label{fig:OII_result_a}
\end{figure*}

\subsection{DIBE$_{Dis}$ Loss can adjust output imbalance}
\label{subsec:DIBEDis_Loss_can_adjust_output_imbalance}
The training curves of $OII$ in various configurations are demonstrated in Fig.\ref{fig:OII_result_a}. 
We can find that the adjustment of $\alpha$ of the losses, i.e. DIBE$_{Dis}$ and Tversky, is significant. 
Regardless of networks and datasets, the $OII$ curves locate apart from each other for different $\alpha$.

As $\alpha$ increases from $0.1$ to $0.9$, the $OII$ curves of DIBE$_{Dis}$ Loss are distributed from top to bottom, no matter SegNet or U-Net (Fig.\ref{fig:OII_result_a}).
And its distributed rule is the same as Tversky Loss, demonstrating the close ability to control the output imbalance for DIBE$_{Dis}$ Loss and Tversky Loss. 
This ability to regulate output imbalance is not possessed by Focal Loss.
Our findings indicate that the $OII$ curves of DIBE$_{Dis}$ Loss were all essentially above $0$. 
We attribute this to the fact that DIBE$_{Dis}$ is a distribution-based loss, whose optimization goal is to enhance the classification accuracy of each pixel; thus, the $PA$ will be elevated. 
This implies a lower $FP$, and correspondingly, a lower $x$ in $OII$, leading to $x$ taking on a value within $(0,1)$ and consequently manifesting as $OII > 0$.

We compared the performance of CE Loss, Focal Loss, and DIBE$_{Dis}$ Loss (Table \ref{table:THC448_dis_Loss}) when training on different models with THC448, with $\alpha$ set at 0.3 throughout. 
Our findings showed that the results of DIBE$_{Dis}$ Loss were generally comparable to that of Focal Loss. 
Moreover, if hyperparameter $\alpha$ optimization was conducted, the performance of DIBE$_{Dis}$ Loss would likely surpass Focal Loss. 
Apart from the performance, we emphasize that DIBE$_{Dis}$ Loss has the unique attribute of being able to address output imbalance, which is not shared by CE Loss and Focal Loss.

\begin{table}\footnotesize
    \centering
    \setlength{\tabcolsep}{2pt}
    \caption{Results on THC448 with different distribution-based losses and networks.}
    \begin{tabular}{ccccc}
    \toprule  
    \multirow{2}{*}{Loss} &\multicolumn{4}{c}{$IoU$}\\
    \cline{2-5}
      & SegNet & U-Net & PSPNet & DeepLabv3\\
    \midrule  
    $CE$ & 0.435&0.577&0.372&0.259\\
    $Focal$ & 0.476&0.585&0.252&0.293\\
    $DIBE_{Dis}$ & 0.493&0.585&0.248&0.286\\
    \bottomrule 
    \end{tabular}
    \label{table:THC448_dis_Loss}
\end{table}

\subsection{DIBE Loss performs better}
\label{subsec:DIBE_Loss_shows_better}
Given the consistency between the objectives of region-based losses and the performance measure of $IoU$, the key to improving $IoU$ of compound losses lies in leveraging the inherent region-based losses. 
To this end, we evaluated different compound losses as well as different region-based losses. 
We trained four semantic segmentation models, i.e. SegNet, U-Net, PSPNet, and DeepLabv3, both with region-based loss functions such as Dice Loss, Tversky Loss, and FT Loss, as well as compound loss functions including Combo Loss, EL Loss, HF Loss, and DIBE Loss.
Their respective hyperparameters refer to Section \ref{subsec:Model_and_configuration}.
As shown in Table \ref{table:THC448_dif_Loss}, U-Net outperformed the other models across all eight loss functions. 
Furthermore, our experiments indicated that DIBE$_{Reg}$ Loss and DIBE Loss generally yielded better results than FT Loss and HF Loss, respectively.
    

\begin{table}\footnotesize
    \centering
    \setlength{\tabcolsep}{2pt}
    \caption{Results on THC448 with different losses and networks.}
    \begin{tabular}{ccccc}
    \toprule  
    \multirow{2}{*}{Loss} &\multicolumn{4}{c}{$IoU$}\\
    \cline{2-5}
      & SegNet & U-Net & PSPNet & DeepLabv3\\
    \midrule
    $Dice$ & 0.615&\textbf{0.641}$\uparrow$&0.513&0.554\\
    $Tversky$ & 0.607&\textbf{0.635}$\uparrow$&0.550&0.554\\
    $FT$ & 0.607&\textbf{0.639}$\uparrow$&0.505&0.539\\
    $DIBE_{Reg}$ & 0.617&\textbf{0.652}$\uparrow$&0.515&0.565\\
    \midrule  
    $Combo$ & 0.595&\textbf{0.632}$\uparrow$&0.545&0.538\\
    $EL$ & 0.593&\textbf{0.640}$\uparrow$&0.549&0.553\\
    $HF$ & 0.602&\textbf{0.644}$\uparrow$&0.547&0.507\\
    $DIBE$ & 0.600&\textbf{0.646}$\uparrow$&0.558&0.519\\
    \bottomrule 
    \end{tabular}
    \label{table:THC448_dif_Loss}
\end{table}

In order to further compare the segmentation results of DIBE Loss, DIBE$_{Reg}$ Loss, FT Loss and HF Loss, we select datasets with various levels of input imbalance, namely, THC448, Blowhole, Crack and CFD, to train the U-Net model with the four loss functions by employing the optimal hyperparameters.
Table \ref{table:U-Net_dif_Loss} presents the resulting performance, from which it can be observed that DIBE Loss and DIBE${Reg}$ Loss demonstrate superiority over HF Loss and FT Loss in all the four datasets. 
Notably, DIBE$_{Reg}$ Loss significantly improves the $IoU$ when applied to THC448 with extreme imbalance in inputs.
\begin{table}\footnotesize
    \centering
    \setlength{\tabcolsep}{2pt}
    \caption{Results of U-Net with different losses and datasets.}
    \begin{tabular}{ccc|cc}
    \toprule  
    \multirow{2}{*}{Dataset} &\multicolumn{4}{c}{$IoU$}\\
    \cline{2-5}
      & $FT$ & $DIBE_{Reg}$ & $HF$ & $DIBE$\\
    \midrule  
    THC448  & 0.639&\textbf{0.652}$\uparrow$&0.644&\textbf{0.646}$\uparrow$\\
    Blowhole & 0.774&\textbf{0.776}$\uparrow$&0.776&\textbf{0.779}$\uparrow$\\
    Crack & 0.708&\textbf{0.717}$\uparrow$&0.716&\textbf{0.723}$\uparrow$\\
    CFD & 0.501&\textbf{0.510}$\uparrow$&0.497&\textbf{0.512}$\uparrow$\\
    \bottomrule 
    \end{tabular}
    \label{table:U-Net_dif_Loss}
\end{table}

\begin{figure}[h]
    \centering
    \includegraphics[width=6cm]{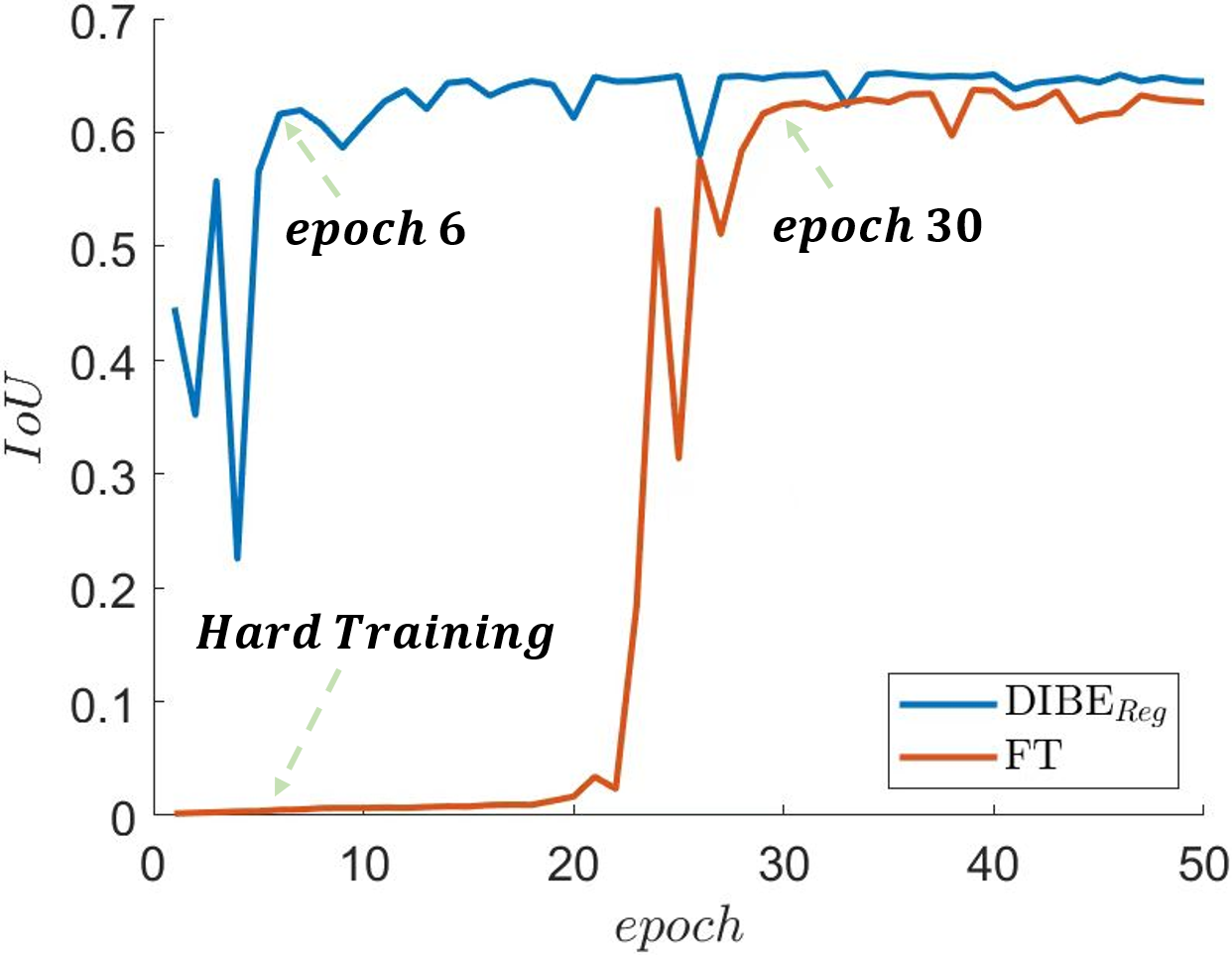}
    
    \caption{Training U-Net under FT Loss and DIBE$_{Reg}$ Loss on THC448 with optimal hyperparameters, DIBE$_{Reg}$ has the advantage of early-stop over FT.}
    \label{fig:EarlyStop}
\end{figure}

Compared to FT Loss, DIBE$_{Reg}$ Loss provides not only improved performance but also simpler training with faster convergence.
The results of training U-Net with FT Loss and DIBE${Reg}$ Loss on THC448, as depicted in Figure \ref{fig:EarlyStop}, highlight the phenomenon of delayed convergence with respect to FT Loss. 
Specifically, its $IoU$ remains almost zero up to the twentieth epoch. 
In contrast, DIBE${Reg}$ Loss achieves a convincing $IoU$ at the sixth epoch. 
This early convergence of DIBE$_{Reg}$ Loss is highly valuable in practical applications since it requires less computing resources and training time compared to FT Loss while still yielding the greater accuracy.
Moreover, its efficient training model will be particularly advantageous when it comes to large-scale hyperparameter optimization.

\section{Discussion}
\label{sec:Discussion}
\textbf{Why not use cropped low imbalanced dataset directly?} 
By cropping images, the number of background pixels can be drastically reduced, thus mitigating the input imbalance issue. 
For instance, experiments had been conducted on THC448, THC224, and THC112 datasets, all of which had been generated by cropping original images and thereby exhibiting gradually reduced input imbalance. 
However, as the cropping degree increase, the inference frequency also increases, causing a dramatic reduction in the inference speed which is unfeasible in a real-world application. 
For this reason, the problem of high or even extreme input imbalance need to be tackled through alternate means (e.g. employing specific models or loss functions).

\textbf{Suggestions on the Selection of the proposed Loss Function:} 
Our experimentation revealed that for extremely-imbalanced input datasets, training with DIBE${Reg}$ Loss resulted in improved performance compared to that of DIBE Loss. 
This may be attributed to the weighting between DIBE${Dis}$ and DIBE$_{Reg}$ in the DIBE Loss used in the experiment, where $\theta=0.5$.
Consequently, we suggest the use of $\theta=0.5$ for mediumly and highly imbalanced application scenarios, while a slightly smaller $\theta$ (such as $\theta=0.3$) should be employed for extremely-imbalanced input datasets.

\textbf{Why $OII$ can guide the selection of hyperparameters?} 
For segmentation tasks, $IoU$ is the most important indicator to measure the quality of the model. 
Nonetheless, for some specific tasks, $PA$ needs to be concerned in addition to $IoU$.
Examining Eqs.\ref{equ:IoU} and \ref{equ:PA}, it can be inferred that the correlation between $IoU$ and $PA$ is governed by the relative size of $FP$ and $FN$, which also determines the degree of output imbalance.
Hence, using $OII$ guidance to select the appropriate hyperparameter to adjust the output imbalance can improve $IoU$ or increase $PA$.

\textbf{Design of output imbalance indicator:} 
We believe the design of $OII$ is still an open topic and a good metric could indicate a direction for research. 
The regulating factor $k$ used in $OII$ is a linear function of $IoU$ and other metrics that can be used to weigh $TP$ against $FP+FN$ are also worth considering.

\section{Conclusion and Future work}
\label{sec:Conclusion_and_Future_work}
This study provices a PV module cells data THC with extreme input and explores a more accurate defect segmentation of monocrystalline PV module cells images from the perspective of loss function, and apply it to the solar energy industry.
For THC segmentation with extreme input imbalance, we proposes three innovative work on both input and output imbalance within the framework of DIBE. 
Firstly, for the unreasonable indicators to measure the output imbalance, a new measure $OII$ is proposed to quantify the output imbalance. 
Secondly, for the distribution-based loss without the ability to adjust output imbalance, a distribution-based DIBE$_{Dis}$ loss is proposed. 
Finally, for the inconsistency between training and inference and the lack of loss against extreme input imbalance, a DIBE Loss, which has an equivalent amount of hyperparameters to HF loss and adapts to extremely-imbalanced input data, is proposed.
The analysis of the loss and its gradient reveals the efficacy of the DIBE Loss, and the proposed method is further verified through exhaustive experiments.

Our method is not only applicable to THC of PV module cells, but also can be applied to other extreme input imbalance scenarios
In the future, we intend to delve further into the correlation between input and output imbalances.
Furthermore, we are willing to study the relationship between each hyperparameter in DIBE Loss, then fuse the hyperparameters with similar functions.
Considering the fact that $OII$ and DIBE Loss have been designed initially for binary classification tasks in semantic segmentation, their potential can further be explored to support multi-class segmentation scenarios.
This highlights the universality of these metrics, enabling them to be applied to a wide range of supervised semantic segmentation tasks.

\section*{Acknowledgement}
\label{sec:Acknowledgement}
This work was supported in part by Grants of National Key R\&D Program of China 2020AAA0108300, in part by the National Nature Science Foundation of China under Grant 62171288, and in part by the Shenzhen Science and Technology Program under Grant JCYJ20190808143415801. 

{\footnotesize
\bibliographystyle{acm}
\bibliography{ref}
}

\clearpage
\begin{appendices}
\newpage
\section{Proof of Proposition 1}
\label{appendixA}
\noindent We ﬁrst restate the proposition.

\noindent\textbf{Proposition 1:} $OII$ is symmetrical, directional, and numerical normalized. That is $OII$ (Eqs.\ref{equ:IoU}, \ref{equ:OII_x} and \ref{equ:OII_K}) satisfies constraint condition (Eqs.\ref{equ:definition1} and \ref{equ:definition2}).

\noindent\textbf{Proof that $OII$ is symmetrical, directional:} 

\noindent Proof.
If $x \in (0, 1]$, then $\frac{1}{x} \in [1, \infty)$ and vice versa.
By introducing $\frac{1}{x}$ into the definition of $OII$ (Eq.\ref{equ:OII}), it can be obtained
\begin{equation}
 OII\Big(\dfrac{1}{x}\Big) =\left\{
\begin{array}{rcl}
k\Big(1-\dfrac{1}{x+\frac{1}{x}-1}\Big), & {\dfrac{1}{x} \in (0,1]},\\
k\Big(\dfrac{1}{x+\frac{1}{x}-1}-1\Big), & {\dfrac{1}{x} \in (1,\infty)}.
\end{array} \right.     
\nonumber\end{equation}

\noindent After simplification, the following can be obtained
\begin{equation}
 OII\Big(\dfrac{1}{x}\Big) =\left\{
\begin{array}{rcl}
-k\Big(1-\dfrac{1}{x+\frac{1}{x}-1}\Big), & {x \in (0,1)},\\
-k\Big(\dfrac{1}{x+\frac{1}{x}-1}-1\Big), & {x \in [1,\infty)}.
\end{array} \right.     
\nonumber\end{equation}

\noindent Comparing the definition of $OII$ with the above formula, it can be concluded that
\begin{equation}
 OII\Big(\dfrac{1}{x}\Big) = - OII(x), \quad {x \neq 1}.
\nonumber\end{equation}

\noindent And because
\begin{equation}
 OII\Big(\dfrac{1}{x}\Big) = 0, \quad {x = 1},
\nonumber\end{equation}
\begin{equation}
 OII(x) = 0, \quad {x = 1},
\nonumber\end{equation}

\noindent we can find that
\begin{equation}
 OII\Big(\dfrac{1}{x}\Big) = -OII(x) = 0, \quad {x = 1}.
\nonumber\end{equation}

\noindent In summary, we have demonstrated that $OII$ is symmetrical, directional:
\begin{equation}
 OII(x) = -OII\Big(\dfrac{1}{x}\Big), \quad {x \in (0, \infty)}
\nonumber\end{equation}

\noindent that is
\begin{equation}
 OII\Big(\dfrac{FP}{FN}\Big) = -OII\Big(\dfrac{FN}{FP}\Big)
\nonumber\end{equation}

\noindent\textbf{Proof that $OII$ is numerical normalized:} 

\noindent Proof.
Let $x$ be an element of $(0, 1]$, the $OII$ is expressed as
\begin{equation}
 OII(x) = k\Big(1-\dfrac{1}{x+\frac{1}{x}-1}\Big).
\nonumber\end{equation}

\noindent Because
\begin{equation}
    x+\frac{1}{x} \geq 2,
\nonumber\end{equation}

\begin{equation}
    0 < \dfrac{1}{x+\frac{1}{x}-1} \leq 1,
\nonumber\end{equation}

\begin{equation}
    1 > 1-\dfrac{1}{x+\frac{1}{x}-1} \geq 0, 
\nonumber\end{equation}

\noindent and
\begin{equation}
    0 \leq IoU \leq 1,
\nonumber\end{equation}

\begin{equation}
    1 \geq k = 1-\dfrac{IoU}{2} \geq 0.5,
\nonumber\end{equation}

\noindent we can conclude that the value range of $OII$ is as follows:
\begin{equation}
 0 \leq OII(x) = k\Big(1-\dfrac{1}{x+\frac{1}{x}-1}\Big) < 1.
\nonumber\end{equation}

\noindent Let $x \in (1, \infty)$ , and similarly can be obtained:
\begin{equation}
 -1 < OII(x) = k\Big(\dfrac{1}{x+\frac{1}{x}-1}-1\Big) \leq 0.
\nonumber\end{equation}

\noindent In summary, we have proved that $OII$ is numerical normalized,
\begin{equation}
 -1 < OII(x) < 1,
\nonumber\end{equation}

\noindent that is
\begin{equation}
 OII\Big(\dfrac{FP}{FN}\Big) \in (-1, 1)
\nonumber\end{equation}

\clearpage
\section{Region-based DIBE Loss}
\label{appendixB}
\begin{figure}
  \centering
  \begin{subfigure}{1\linewidth}
    \centering
    \includegraphics[width=6cm]{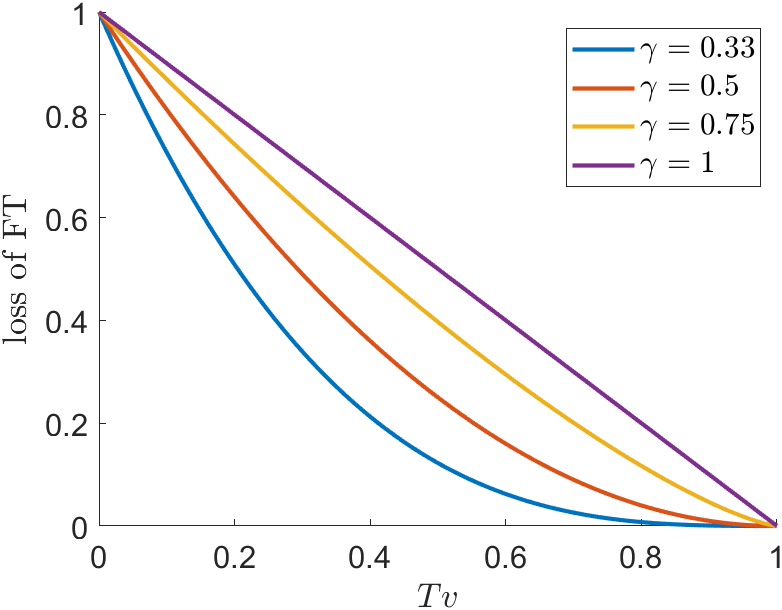}
    \caption{Focal Tversky Loss}
    \label{fig:Loss_FT}
  \end{subfigure}
  \hfill
  \begin{subfigure}{1\linewidth}
    \centering
    \includegraphics[width=6cm]{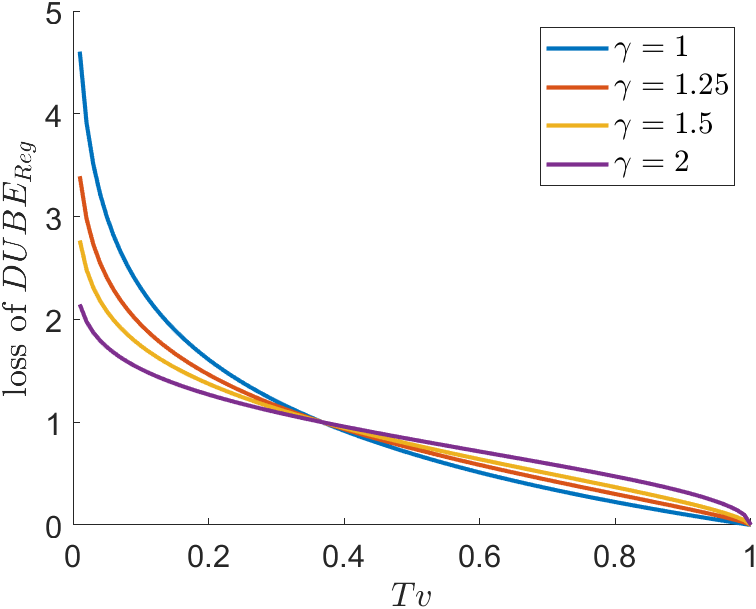}
    \caption{DIBE$_{Reg}$ Loss}
    \label{fig:Loss_DIBERe}
  \end{subfigure}
  \caption{(a) Curve of FT Loss changing with Tversky coefficient. (b) Curve of DIBE$_{Reg}$ Loss changing with Tversky coefficient.}
  \label{fig:Loss_Region}
\end{figure}

Comparing the curve of DIBE$_{Reg}$ Loss versus $Tv$ with that of FT Loss versus $Tv$, they are shown in Fig.\ref{fig:Loss_Region}. 
It can be found that in the region close to $0$, DIBE$_{Reg}$ Loss is always greater and decreases faster than FT Loss. While $Tv\rightarrow 1$, DIBE$_{Reg}$ Loss not only has a larger gradient magnitude, but also will not give rise to over-suppression.

In order to further compare in the scene of extremely-imbalanced input, we take the derivatives of Eqs.\ref{equ:FT} and \ref{equ:DIBERe} respectively as follows.
\begin{equation}
\frac{\partial \mathcal{L}_{FT}}{\partial Tv} = -\frac{1}{\gamma}(1-Tv)^{\tfrac{1}{\gamma}-1},
\label{equ:dFT}
\end{equation}
\begin{equation}
\frac{\partial \mathcal{L}_{DIBE_{Reg}}}{\partial Tv} = -\frac{1}{\gamma Tv}(-\log(Tv))^{\tfrac{1}{\gamma}-1}.
\label{equ:dDIBERe}
\end{equation}

\begin{figure}
  \centering
  \begin{subfigure}{1\linewidth}
    \centering
    \includegraphics[width=6cm]{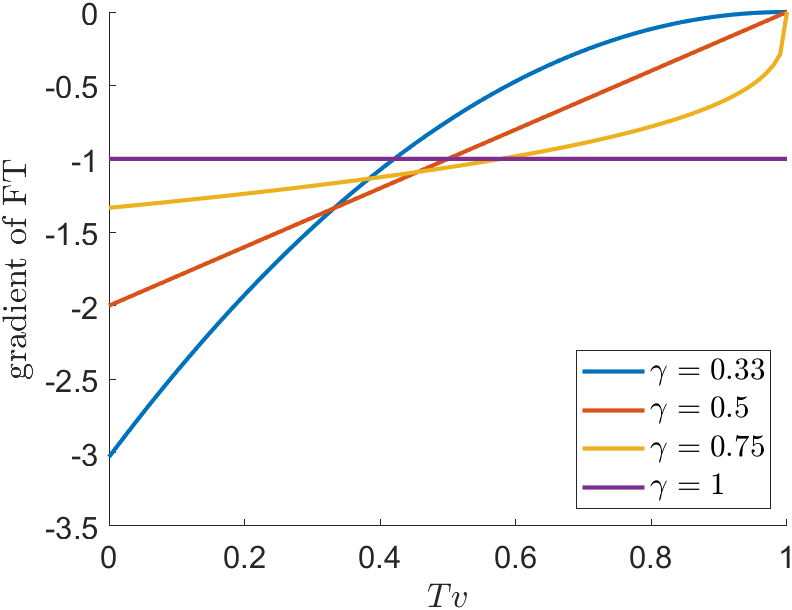}
    \caption{Gradient of Focal Tversky Loss}
    \label{fig:Gradient_FT}
  \end{subfigure}
  \hfill
  \begin{subfigure}{1\linewidth}
    \centering
    \includegraphics[width=6cm]{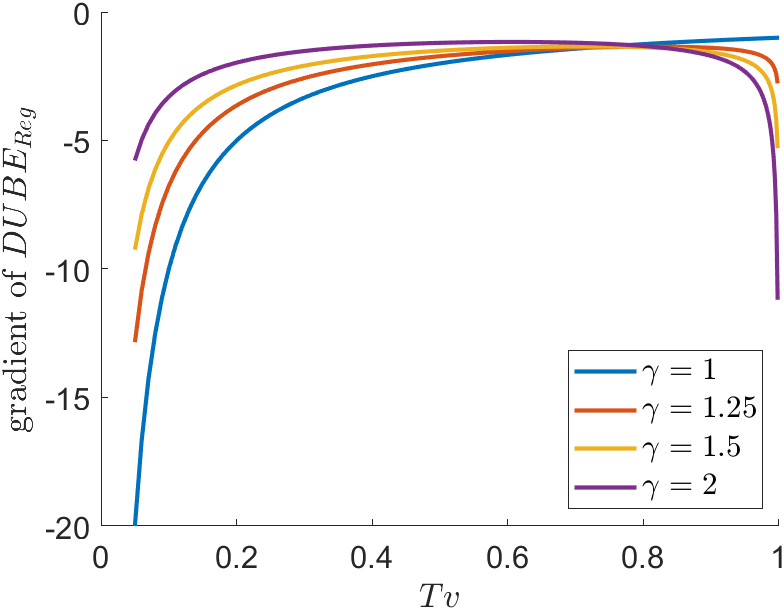}
    \caption{Gradient of DIBE$_{Reg}$ Loss}
    \label{fig:Gradient_DIBERe}
  \end{subfigure}
  \caption{Gradients of FT Loss and DIBE$_{Reg}$ Loss versus Tversky coefficient respectively.}
  \label{fig:Gradient_Region}
\end{figure}

The gradients versus $Tv$ are plotted in Fig.\ref{fig:Gradient_Region}. 
By observing Fig.\ref{fig:Gradient_FT}, it is found that in the region close to 0, the gradient magnitude of FT Loss is larger than that of Tversky Loss ($\gamma=1$ in FT Loss). 
With the increase of $Tv$, however, the gradient magnitude of FT Loss continues to decrease until to zero when $Tv$ approaches $1$. 
This is exactly the over-suppression phenomenon of FT loss. By contrast, Fig.\ref{fig:Gradient_DIBERe} shows that the gradient magnitude of DIBE$_{Reg}$ Loss is much larger than that of FT Loss in the neighbor of $Tv=0$. 
As $Tv$ increases, the gradient magnitude of DIBE$_{Reg}$ Loss decreases at the beginning and then increases rapidly in the neighbor of $1$. 
This not only settles the problem of over-suppression, but also helps DIBE$_{Reg}$ Loss focus on small targets segmentation.

\begin{figure}[h]
    \centering
    \includegraphics[width=6cm]{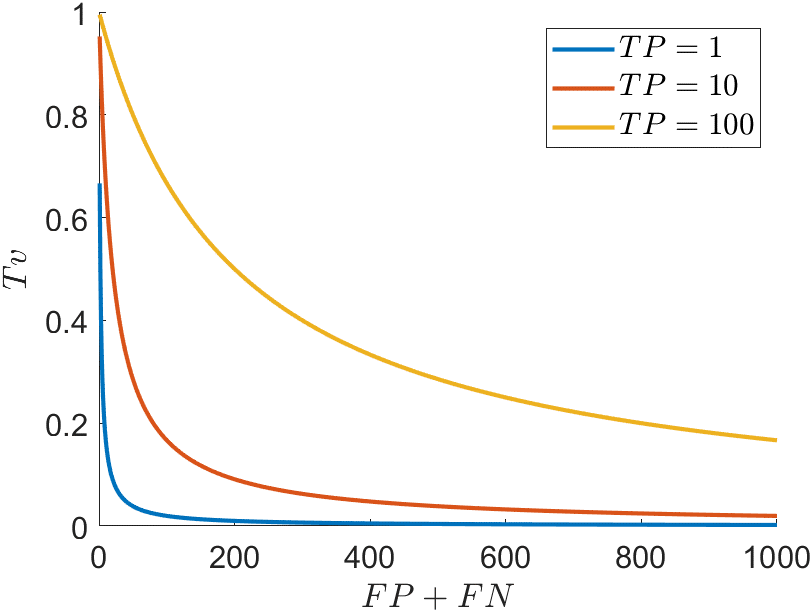}
    
    \caption{Curve of $Tv$ regarding $TP$, $FP$ and $FN$.}
    \label{fig:difTP_Tv}
\end{figure}

In the training phase of network, $FP$ and $FN$ are decrease and $TP$ increases. 
The degree of input imbalance has a great impact on the relationship of those three.
The changes of the three further determine the changes of $Tv$ (see Eq.\ref{equ:Tversky}). 
To study the effect of input imbalance on $Tv$, we set $\alpha=\beta=0.5$ and treat $FP$ and $FN$ as a whole. 
Considering that the higher the level of input imbalance, the smaller $TP$ when $FP$ and $FN$ are fixed, $TP$ can thus vary to simulate the change of $Tv$ in different levels of input imbalance. 
We use, for example, $TP=1,10,100$ in Fig.\ref{fig:difTP_Tv} respectively to represent the extreme, high and normal level input imbalance. 
For Fig.\ref{fig:difTP_Tv}, it can be recognized that with the increase of input imbalance, $Tv$ will stay close to $0$ for a long time during training. 
Referring to the loss plot (Fig.\ref{fig:Loss_Region}) and the gradient plot (Fig.\ref{fig:Gradient_Region}) of DIBE$_{Reg}$ Loss, both the loss and gradient magnitude value of DIBE$_{Reg}$ Loss are larger than that of FT Loss when $Tv\rightarrow 0$.
Therefore DIBE$_{Reg}$ Loss has more advantages in dealing with the extremely-imbalanced input data.

\begin{figure*}[h]
    \centering
    \includegraphics[width=16cm]{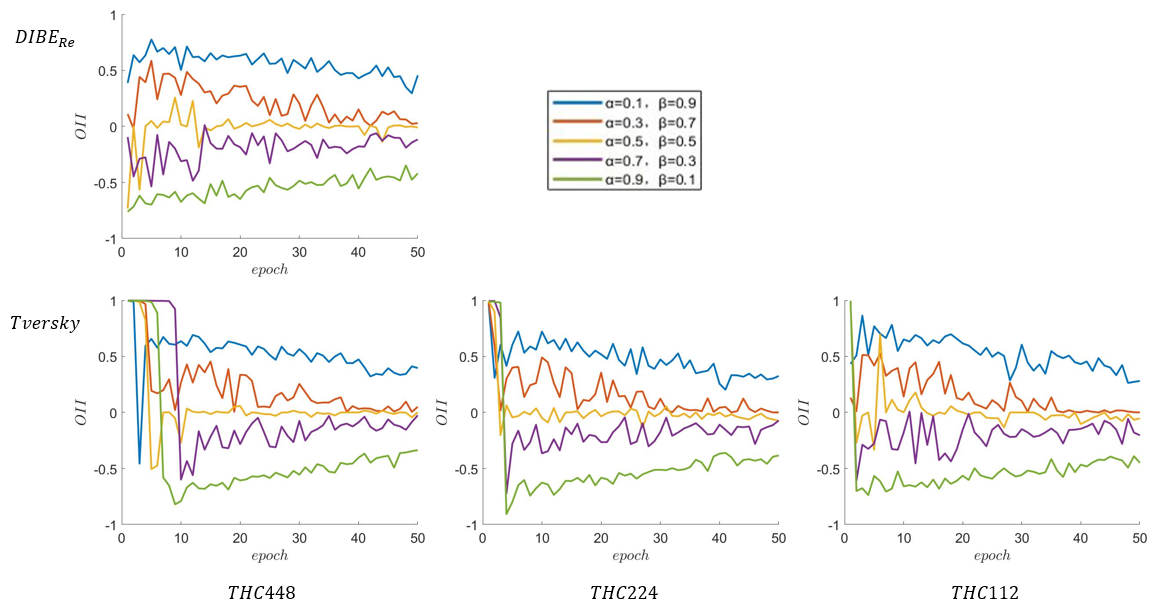}
    
    \caption{Regulation of output imbalance of $\alpha$ under different input imbalance (THC112, THC224, THC448) and different losses.}
    \label{fig:OII_result_b}
\end{figure*}

We conducted some experiments on DIBE$_{Reg}$ loss and visualized the experimental results as shown in Fig.\ref{fig:OII_result_b}. 
Observing the $OII$ curves of Tversky in THC112, THC224 and THC448 with U-Net (bottom of figure), 
it can be found that the higher the degree of input imbalance, the more difficult Tversky Loss is to be minimized.

By comparing the results of using Tversky Loss to train THC112 and using DIBE$_{Reg}$ Loss to train THC448, we found that their $OII$ curves are almost the same.
It means that after using logarithmic operation, DIBE$_{Reg}$ Loss is more beneficial to the loss minimization of small targets. 
It can also be understood that the process of DIBE$_{Reg}$ Loss minimization for extremely-imbalanced input is similar to that of Tversky Loss minimization for medium/lightly imbalanced input. 
Therefore, DIBE$_{Reg}$ Loss is more suitable for extremely-imbalanced input data.

\clearpage
\section{$OII$ guidance for hyperparameters selection}
\label{appendixC}
\begin{figure}[h]
    \centering
    \includegraphics[width=5.5cm]{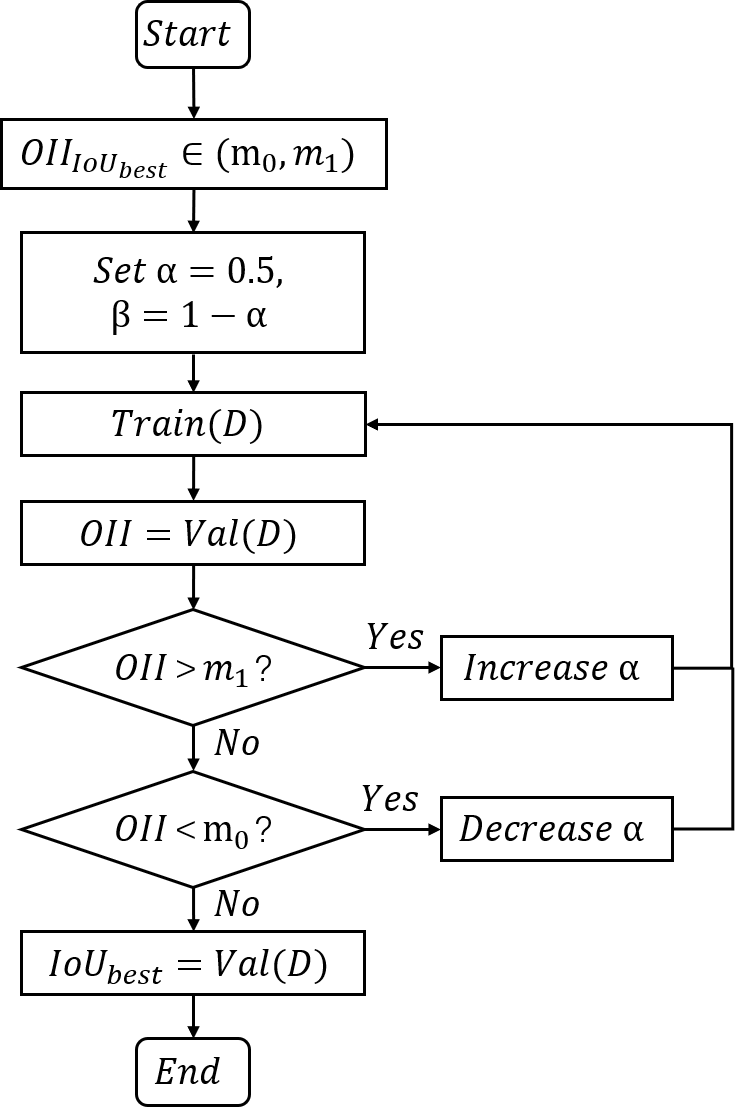}
    
    \caption{Use $OII$ to guide the selection of $\alpha$ and $\beta$.}
    \label{fig:FlowChart}
\end{figure}

$OII$ can also be used to guide the selection of hyperparameters $\alpha$ and $\beta$ to adjust the output imbalance. The algorithm flowchart (Fig.\ref{fig:FlowChart}) shows how to use $OII$ guidance to find the best $\alpha$ and $\beta$.
Firstly, when the $OII$ value of the input data $D$ falls within the range of $(m_0, m_1)$, we believe the performance of the model can be enhanced. 
In this research, the THC448 dataset was analyzed and it was established that $IoU$ and $PA$ values could be improved upon when $OII \in (0.15,0.25)$. 
Following this, a model was trained and $OII$ was computed on the validating dataset by setting $\alpha=\beta=0.5$. 
Subsequently, the value of $OII$ was evaluated, and the selection of the hyperparameters $\alpha$ and $\beta$ was readjusted using the Bisection method. 
For example, in this experiment, if the $OII$ is greater than $0.25$, increase $\alpha$ appropriately and train again to get a better $IoU$ or $PA$. 
Eventually, by repeatedly readjusting $\alpha$ and $\beta$, the $OII$ could be brought in the target area $(m_0, m_1)$ and the model performance on $IoU$ or $PA$ could be substantially improved.
In order to verify the hyperparameters guidance of $OII$, we put more detailed experimental analysis in Subsection \ref{subsec:OII_can_reflect_and_guide_output_imbalance}.

At present, the most commonly used hyperparameters optimization method is grid search. 
The superiority of $OII$ guidance over grid search is that it greatly saves computational time and improves the efficiency of hyperparameters optimization. 
In general, we compare the two as follows.
If grid search is used to catch the $\alpha$ in interval $[a, b]$ which is divided into $k$ equal part, we set $\alpha_{0}=a$, $\alpha_{1}=a+(b-a)/k$, $\alpha_{2}=a+2(b-a)/k$, ..., $\alpha_{k-1}=a+(k-1)(b-a)/k$, $\alpha_{k}=b$ to train $k+1$ models. 
If using $OII$ guidance to optimise $\alpha$ in the same interval $[a, b]$ which then is divided into $2^n$ equal part, we need to constantly adjust $\alpha$ with Bisection method to make $OII$ fall into the target interval $(m_0, m_1)$. 
It means that we will select $\alpha_{0}=a+(b-a)/2$, $\alpha_{1}=a+(b-a)/2^2$, $\alpha_{2}=a+(b-a)/2^3$, ..., $\alpha_{n-1}=a+(b-a)/2^n$ to train $n$ models (with the optimal choice of $\alpha$ is $a+(b-a)/2^n$ as an example). 
In fact, when $\alpha_{s}=a+(b-a)/2^{s+1}$, where $0\leqslant s\leqslant n-1$, the $OII$ may fall into $(m_0, m_1)$, so no more than $n$ models will be trained. 
In order to fairly compare the complexity of the two hyperparameters optimization methods, we maintain the same refinement of them, dividing interval $[a, b]$ into $2^m$ equal part. 
The length of each subinterval is $(b-a)/2^m$. 
In this case, using grid search for optimization requires training $2^m+1$ models. 
If $OII$ guidance is used for optimization, the number of models to be trained should be no more than $m$. 
This means that the optimization efficiency of $OII$ guidance is more than $(2^{m}+1)/m$ times that of grid search (e.g., the optimization efficiency of the $OII$ guidance is more than 100 times that of grid search when $n=10$).

\begin{table}\footnotesize
    \centering
    \setlength{\tabcolsep}{2pt}
    \caption{Grid Search and $OII$ guidance of $\alpha$ in DIBE$_{Dis}$, $Tversky$ and DIBE$_{Re}$ Loss with SegNet and U-Net.}
    \begin{tabular}{cccccccccc}
    \toprule  
    &\multirow{2}{*}{Loss}& \multicolumn{4}{c}{SegNet}& \multicolumn{4}{c}{U-Net}\\
    \cline{3-10}
    && times & $\alpha$ & IoU & PA& times & $\alpha$ & IoU & PA\\
    \midrule
    \multirow{3}{*}{Basic}&DIBE$_{Dis}$&1 &0.5 &0.442 &0.713&1 &0.5 &0.581 &0.832\\
    &$Tversky$&1 &0.5 &0.620 &0.774 &1 &0.5 &0.642 &0.782\\
    &DIBE$_{Re}$&1 &0.5 &0.625 &0.797 &1 &0.5 &0.644 &0.794 \\
    
     \midrule
    \multirow{2}{*}{Grid}&DIBE$_{Dis}$&5 &0.1 &0.481 &0.823 &5 &0.7 &0.586 &0.800\\
    \multirow{2}{*}{Search}&$Tversky$&5 &0.5 &0.620 &0.774 &5 &0.5 &0.642 &0.782\\
    &DIBE$_{Re}$&5 &0.5 &0.625 &0.797 &5 &0.5 &0.644 &0.794\\
    \midrule  
    \multirow{2}{*}{$OII$}&DIBE$_{Dis}$&2$\downarrow$ &0.7 &0.477 &0.709 &2$\downarrow$ &0.7 &0.586 &0.800\\
    \multirow{2}{*}{$Guide$}&$Tversky$&2$\downarrow$ &0.3 &0.619 &0.830 &3$\downarrow$ &0.1 &0.635 &0.863\\
    &DIBE$_{Re}$&2$\downarrow$ &0.3 &0.624 &0.834 &3$\downarrow$ &0.1 &0.632 &0.852\\
    \bottomrule 
    \end{tabular}
    \label{table:gridsearch_guidance_Hyperparameters}
\end{table}
We believe the goal of $OII$ of THC448 is in the interval of $(0.15, 0.25)$.
According to the guidance algorithm mentioned above and Bisection method, we use $OII$ guidance to select hyperparameter $\alpha\in [0.1,0.9]$ under DIBE$_{Dis}$, Tversky and DIBE$_{Re}$ Loss and record $IoU$, $PA$ and training times in Table \ref{table:gridsearch_guidance_Hyperparameters}.

It is recognized that the method of using $OII$ to guide the selection of hyperparameters is completely applicable. 
Comparing the basic and $OII$ guidance in the Table \ref{table:gridsearch_guidance_Hyperparameters}, it can be found that using $OII$ guidance can significantly boost $IoU$ or $PA$.
By comparing the optimal results found by $OII$ guidance and grid search in Table \ref{table:gridsearch_guidance_Hyperparameters}, it can be observed that in the majority of scenarios, $OII$ guidance yields performance comparable to that of grid search while requiring less training times.
If the step size of $\alpha$ is reduced to make the search more precise, the number of training times of $OII$ guidance compared to grid search will be significantly reduced.


Two examples of OII guiding hyperparameters selection are as follows:

1. Use DIBE$_{Dis}$ Loss to train SegNet. 
At the beginning, $\alpha_0=\beta=0.5$ (centroid of $[0.1, 0.9]$) is set to train the network and yields $IoU=0.442$, $PA=0.713$, $OII=0.294$. 
$\alpha$ is then increased to reduce $OII$ because $OII$ is greater than $0.25$ at this time. 
Then we set $\alpha_1=0.7$ (centroid of $[\alpha_0, 0.9]$) to train the model again and obtain $IoU=0.477$, $PA=0.709$, $OII=0.162$. 
After adjusting $\alpha$, $OII$ has obviously been decreased to fall into the goal interval $(0.15, 0.25)$. Thereby we successfully trade small $PA$ loss ($-0.56\%$) for greatly improved $IoU$ ($+7.92\%$).

2. Use Tversky Loss to train U-Net. 
At the beginning, $\alpha_0=\beta=0.5$ is set to train the network and yields $IoU=0.642$, $PA=0.782$, $OII=0.000$. 
$\alpha$ is then reduced to increase $OII$ because $OII$ is less than $0.15$ (centroid of $[0.1, \alpha_0]$) at this time.
Then we set $\alpha_1=0.3$ to train the model again and obtain $IoU=0.642$, $PA=0.798$, $OII=0.022$. 
After that, $OII$ is still less than $0.15$, and we continue to decrease $\alpha$ and set $\alpha_2=0.2$ (centroid of $[0.1, \alpha_1]$) to train the model again and it yields $IoU=0.636$, $PA=0.840$, $OII=0.228$. 
It can be found that $OII$ is then tuned to fall into the goad interval and $PA$ has been greatly improved ($+7.4\%$) when $IoU$ is almost unchanged ($-1.2\%$). 

\end{appendices}

\end{document}